\documentclass[runningheads]{llncs}

\usepackage{eccv}

\usepackage{eccvabbrv}

\usepackage[pagebackref]{hyperref}
\usepackage{graphicx}
\usepackage{booktabs}
\usepackage{multirow}
\usepackage{array}
\usepackage[table]{xcolor}
\usepackage{tabularx}
\usepackage{makecell}      %
\usepackage{xcolor}        %
\usepackage{enumitem}
\usepackage{tcolorbox}
\tcbuselibrary{skins,breakable}
\usepackage[accsupp]{axessibility}  %
\usepackage{subcaption} %
\usepackage{adjustbox}
\usepackage{array} %
\usepackage{caption}    %
\usepackage{adjustbox}  %

\usepackage{hyperref}

\usepackage{orcidlink}

\begin{document}

\title{Test-Time Modification: Inverse Domain Transformation for Robust Perception} 

\titlerunning{Test-Time Modification [TTM]}

\author{Arpit Jadon\textsuperscript{1,*} \and
Joshua Niemeijer\textsuperscript{2,*} \and
Yuki M. Asano\textsuperscript{3}}

\authorrunning{A.~Jadon et al.}

\institute{German Aerospace Center Berlin \and German Aerospace Center Braunschweig \and
University of Technology Nuremberg}

\maketitle
\begin{figure}[t]
  \centering
  \includegraphics[width=.535\linewidth]{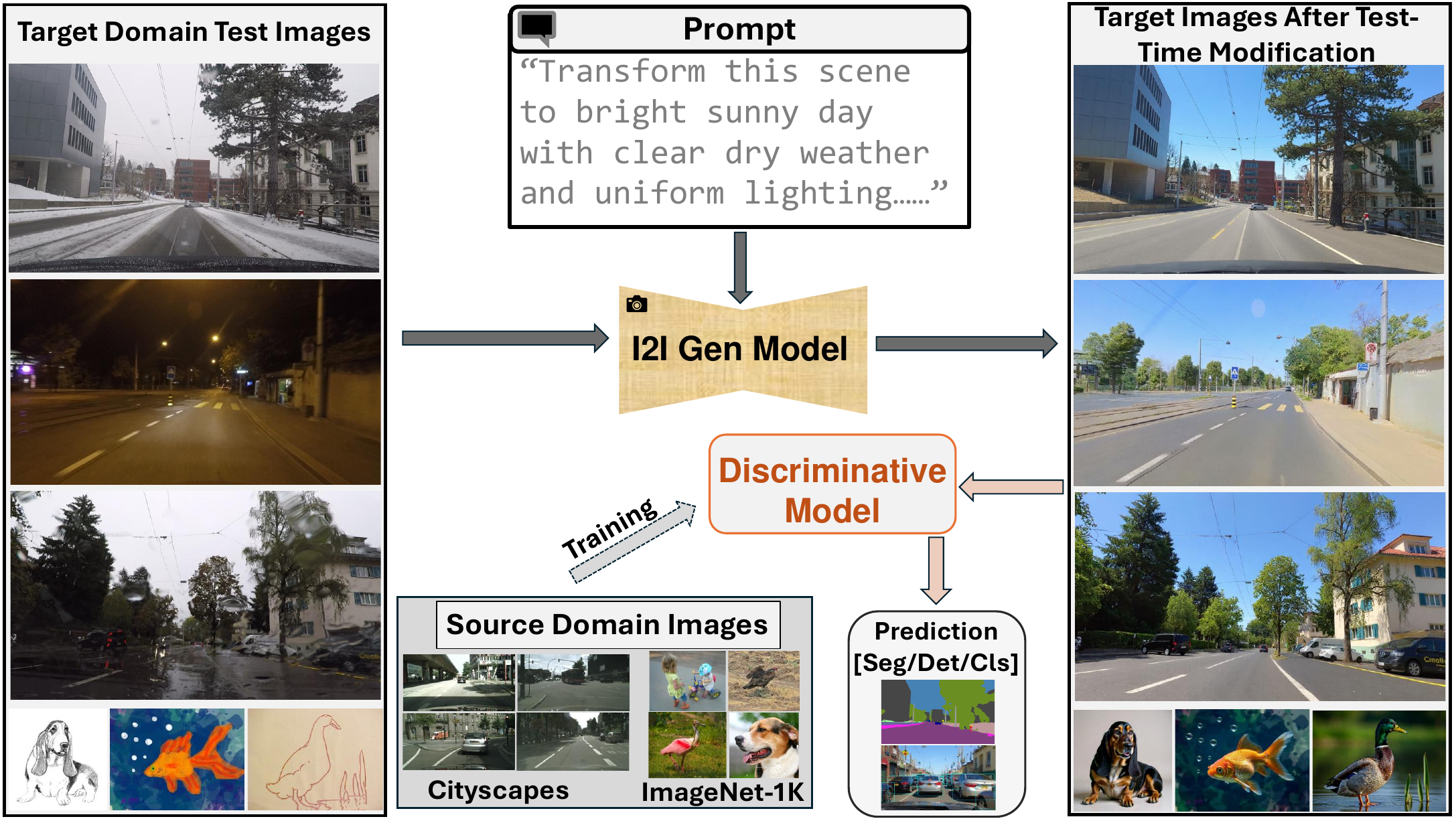}\hfill
  \includegraphics[width=.46\linewidth]{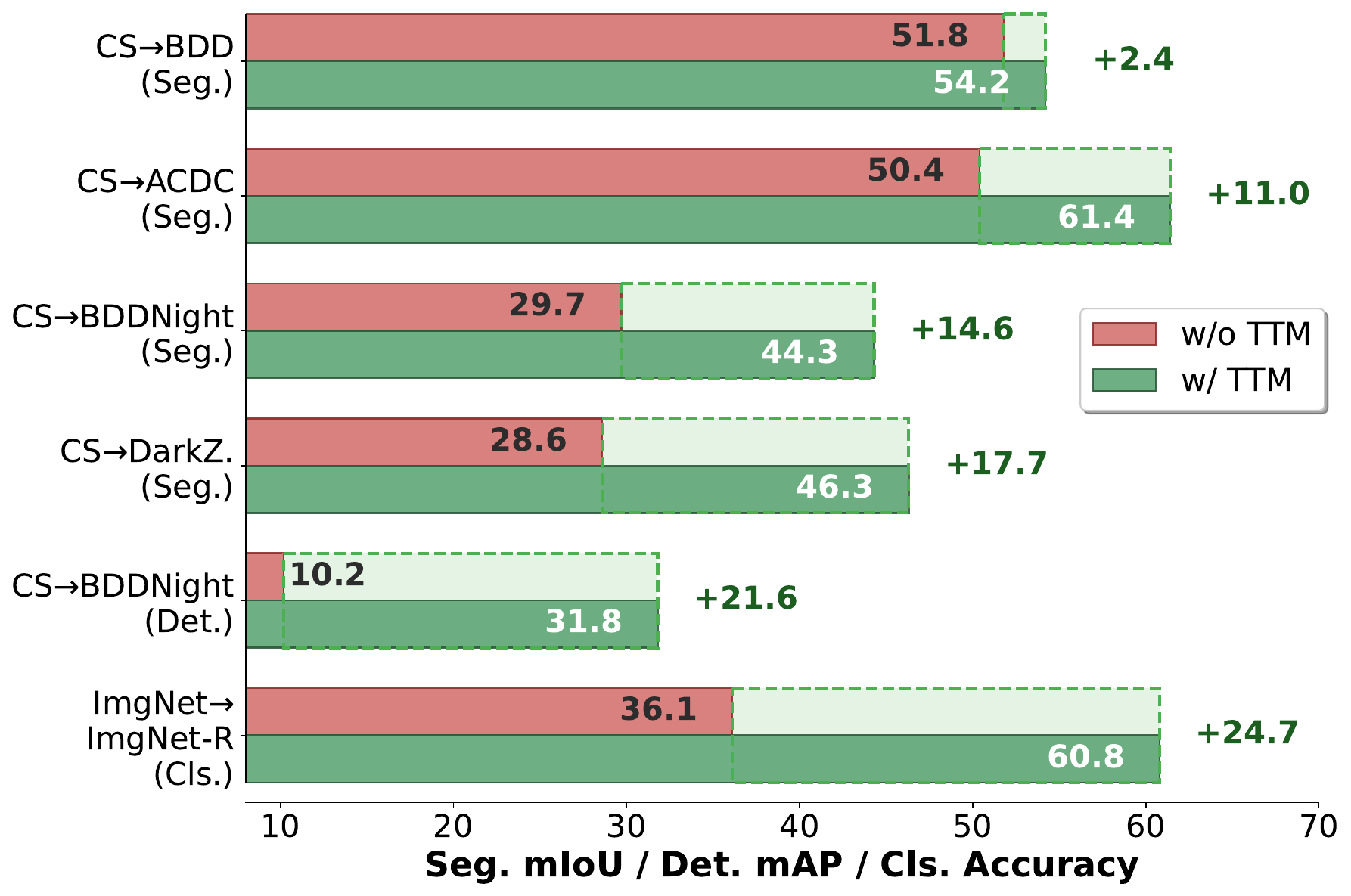}
  \caption{\textbf{Left}: Our proposed Test-time view modification via inverse domain transfer using a strong image-to-image (I2I) Foundation Model. \textbf{Right}: Subset of reported performance improvements on domain generalization benchmarks.}
  \label{fig:teaser_and_summary}
  \vspace{-1.5em}
\end{figure}

{
  \let\thefootnote\relax
  \footnotetext{* Equal contribution}
}

\begin{abstract}
Generative foundation models contain broad visual knowledge and can produce diverse image variations, making them particularly promising for advancing domain generalization tasks. They can be used for training data augmentation, but synthesizing comprehensive target-domain variations remains slow, expensive, and incomplete. We propose an alternative: using diffusion models at test time to map target images back to the source distribution where the downstream model was trained. This approach requires only a source domain description, preserves the task model, and eliminates large-scale synthetic data generation.
We demonstrate consistent improvements across segmentation, detection, and classification tasks under challenging environmental shifts in real-to-real domain generalization scenarios with unknown target distributions. Our analysis spans multiple generative and downstream models, including an ensemble variant for enhanced robustness. 
The method improves BDD100K-Night-Det mAP@50 from 10.2 to 31.8, ImageNet-R top-1 from 36.1 to 60.8, and DarkZurich mIoU from 28.6 to 46.3.

\vspace{-0.5em}
\end{abstract}

\section{Introduction}
\label{sec:intro}
\begin{figure*}[t]
  \centering
  \includegraphics[width=0.99\textwidth]{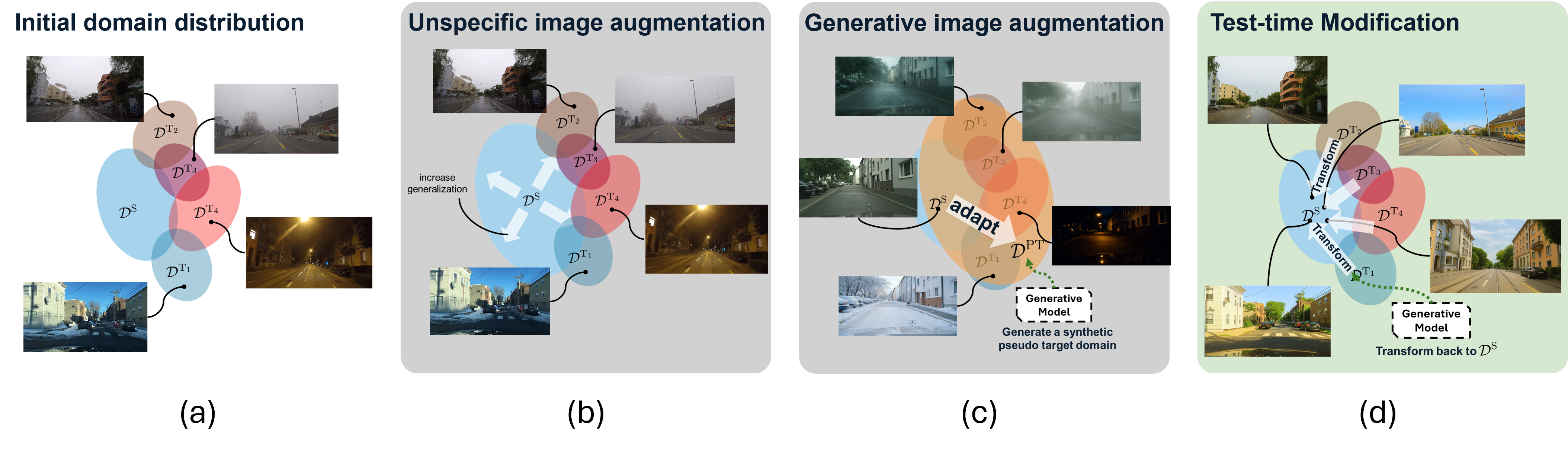}
  \caption{(a) Initial distributions: $\mathcal{D}^S$ source domain, $\mathcal{D}^{T_{k}}$ different target domain distributions. (b) untargeted extension of the source domain by traditional augmentations. (c) use of generative models to create pseudo target domain that compasses all relevant $\mathcal{D}^{T_{k}}$. (d) use of generative models to transform images from $\mathcal{D}^{T_{k}}$ back to the distribution of $\mathcal{D}^{S}$.   
  }
  \label{fig:wide}
  \vspace{-1.5em}
\end{figure*}

Despite major advances in vision architectures and training strategies, model performance still relies strongly on the quality and coverage of the training data. Predictions tend to be reliable when test images follow the same distribution as the training set, but even moderate distribution shifts can lead to a clear drop in accuracy.
To tackle this, the field of domain generalization aims to train models that remain robust to unseen target domains without assuming access to data or annotations from those domains.
Most existing approaches fall into two categories (\cref{fig:wide}b,c).

The first one ``Unspecific Image Augmentation'' modifies the training data in broad, unspecific ways: typically through heavy augmentations such as blur, noise, or color jitter, with the hope that the model will generalize to new domains.
The second approach, ``Generative Training Image Augmentation'' leverages recent diffusion-based generative models to synthesize additional training data. Using text prompts, these methods attempt to approximate potential target domains and generate corresponding images. 
Estimating all relevant target domains 
is difficult, and the diversity of synthesized data remains limited.

In this work, we introduce a third direction.
Instead of expanding the training distribution, we use modern generative models to compute an \textit{inverse transformation at test time}. The goal is to map a target-domain image back towards the source-domain distribution on which the discriminative model was trained. 
This idea is related to test-time normalization techniques such as adjusting BatchNorm statistics, but diffusion-based transformations allow for far richer and more expressive changes. 
\textit{Unlike test-time adaptation (TTA), our approach does not require target-domain  statistics.}
Because the discriminative model performs best on its original training distribution, this inverse mapping improves prediction quality without modifying the model itself.
Importantly, computing this inverse transformation requires only a description of the source domain, which is readily available. In contrast, expanding the training distribution with synthetic data requires anticipating all relevant target domains, which is a far harder problem. Moreover, compared to retraining-based domain adaptation pipelines, our method trades expensive offline retraining for a lightweight inference-time transformation. 

Our approach introduces a single image-to-image transformation step at inference time, implemented using modern generative models. Recent advances in efficient architectures and optimized inference enable this step to run in near real time while preserving accuracy (\cref{para:deployment}).Because the transformation is independent of the downstream model, improvements in generative model efficiency directly translate to faster deployment.
Our method showcases a new paradigm where generative models serve as powerful domain translators, unlocking their learned representations to bridge distribution gaps without any task-specific training. This opens new research directions for understanding and utilizing the complementary strengths of generative and discriminative models. %

We evaluate our inverse domain transformation on domain generalization benchmarks for semantic segmentation, detection, and classification. Our test-time modification  (TTM) pipeline improves performance across all settings without finetuning either the discriminative or generative models, offering a simple, task-agnostic approach to domain generalization.
Our main contributions are:
\begin{itemize}
    \item A formalisation of and recipe for Inverse Domain Transformation for view modification of test data
    \item Quantitative evaluation and analysis of Test-Time Modification (TTM) as a new paradigm that requires no retraining or fine-tuning
    \item State-of-the-art results for various pretrained models on benchmarks such as BDD~\cite{yu2020bdd100k}, DarkZurich~\cite{sakaridis2019guided}, ACDC~\cite{sakaridis2021acdc}, ImageNet-R~\cite{hendrycks2021many}
\end{itemize}

\section{Related Work}
\label{sec:related_work}

Domain Generalization is the task of making a discriminative model robust to data from distributions unseen during training. 
Given a discriminative model trained on a so-called source domain, it is tested on data from unseen target domains that exhibit substantial domain shifts with respect to, for instance, lighting, weather, or environmental conditions.
We present existing approaches to domain generalization and works related to our inverse transformation approach. %

\textbf{Constraining the distributions.}
Early methods constrain input distributions via normalization in latent space, using batch-norm~\cite{SEGU2023109115} or instance normalization~\cite{Pan2018IBNNet} with test-time parameters from target images. Subsequent works introduce whitening~\cite{pan2019Whitening} to decorrelate channel representations and reduce style information, with various improvements for guided whitening~\cite{Choi2021,xu2022dirl,Peng2022semanticaware}. Our approach instead leverages generative models' world knowledge to map target distributions back to source, providing more robust transformations that handle aleatoric uncertainty.

\textbf{Data augmentation and randomization.}
Extending source distributions to include target domain information has been widely explored, from basic augmentations (Gaussian blur, color jitter) to domain randomization~\cite{Yue2019DomainRandomization} using random style transformations. Several works~\cite{peng2021DomRandGlobal,kim2021DomRandwedge} leverage auxiliary domains (paintings, web images) for style randomization, while Huang et al.~\cite{Huang2021DomRand} perform randomization in frequency space.
Zhong et al.~\cite{Zhong2022advStyle} proposed an adversarial style augmentation and Zhao et al.~\cite{Zhao2022Halluc} a style hallucination module.

\textbf{Prior knowledge for source domain extension.}
Generative models like Stable Diffusion~\cite{rombach2022highresolutionimagesynthesislatent} enable synthetic target domain creation using prior knowledge. Niemeijer et al.\cite{Niemeijer2024WACV} generate pseudo-target domains with weather and traffic variations for autonomous driving, then apply unsupervised domain adaptation~\cite{Schwonberg2023Survey}. 
Benigmim et al.~\cite{Benigmim_2024_CVPR} utilize the prior knowledge contained in LLMs to create meaningful prompts the generation of relevant synthetic data. 
ControlNet variants~\cite{zhang2023adding,li2024adversarial,Niemeijer2025ICCV,Li2024ControlNetIC} enable supervised training on synthetic data by conditioning on both labels and target domain descriptions. However, limited prior knowledge of prevents comprehensive coverage of all relevant domains that might be encountered at test-time.
In contrast, we argue that using these generative models is more efficient when done in reverse: we describe the source domain and transform target images back to the source distribution.

\textbf{Inverse transformations.}
Prior inverse transformation methods use style transfer via Fourier transforms~\cite{Termöhlen2021Continual} or GANs~\cite{mutze2022Iverse}, but require \textit{target domain access} to determine style parameters. 
Our transformation does not require additional information from the given target domain image, distinguishing it from test time adaptation (TTA). We therefore do not compare to TTA approaches in the experiments. Further, while generalist video models can solve perception tasks zero-shot~\cite{wiedemer2025video}, they are computationally expensive and still underperform task-specific discriminative models that TTM is designed to enhance --- by employing generative models as a targeted pre-processor that bridges domain gap at test time with reliable single-pass inference. The approach of Yu et al.~\cite{Yu_2023_CVPR} is closest to ours.  
They proposed an inverse transformation to the source domain based on a source domain trained unconditional diffusion model. 
However, this strategy relies on a source-trained diffusion model and is studied primarily in a classification-only setting. In contrast, we introduce a new paradigm that leverages the world knowledge encoded in off-the-shelf foundation editing models (e.g., Flux1. Kontext and Qwen-Image-Edit-2509) together with a text-defined source description, yielding a domain-agnostic, plug-and-play inverse transformation that generalizes across segmentation, detection, and classification.
Kontext~\cite{labs2025flux1kontextflowmatching} and Qwen-Image-Edit-2509~\cite{wu2025qwenimagetechnicalreport} to perform arbitrary transformations using only text descriptions. Beyond style transfer, these models can interpret image content and even reduce aleatoric uncertainty, as we describe in~\cref{sec:aleatoric}.

\section{Methods}
\label{sec:methods}
\label{sec:method}

\noindent \textbf{Preliminaries.} Consider a model $f_\theta$ (for classification, detection, or segmentation) with parameters $\theta$, pretrained on a source domain distribution $\mathcal{D}^S$. Our goal is to generalize this model to unseen target domains $\mathcal{D}^{T_k}$, where $k \in \{1, \ldots, K\}$ represents $K$ potential target domains.

We denote images as $x_n \in \mathbb{R}^{H \times W \times 3}$ for RGB inputs, where $n \in \{1, \ldots, N\}$ indexes a dataset of $N$ images. The model $f_\theta$ produces predictions $\mathbf{y}_n$ for input $x_n$, with $y_{n,c}$ representing the posterior probability for class $c \in \mathcal{C}$.

Generative models like Stable Diffusion~\cite{rombach2022highresolutionimagesynthesislatent,wu2023datasetdm} are trained on large-scale web data containing diverse image distributions, enabling them to capture extensive world knowledge. Recent works \cite{Niemeijer2024WACV,kupyn2024dataset,wu2023diffumask} have exploited this capability to augment training data distributions. In these approaches, a diffusion generative model $G$ is guided by text prompt $t_k$ to create synthetic images $x^{PT}_n$ that approximate target domain characteristics:
\begin{equation}
    x_n^{PT}= G(x_n^{S},t_k)
    \label{eq::xpt}
\end{equation}

These methods estimate relevant target domains by encoding prior knowledge about $\mathcal{D}^{T_{k}}$ distributions into the text prompt $t_k$, producing a pseudo-target domain $\mathcal{D}^{PT}$.

However, this approach assumes we possess sufficient prior knowledge to generate distributions for all unseen domains $\mathcal{D}^{T_{k}}$. We argue this assumption is impractical: (1) the number of relevant target domains may be unbounded, and (2) symbolic text descriptions may fail to capture the full complexity of these distributions. Therefore, we propose to invert this paradigm.

\begin{tcolorbox}[
  enhanced,breakable,
  title={Definition: Test-Time Modification (TTM)},
  fonttitle=\bfseries,
  colback=gray!3,
  colframe=black,
  boxrule=0.6pt,
  arc=1mm,
  left=1.5mm,right=1.5mm,top=1mm,bottom=1mm,
  before skip=6pt, after skip=6pt,
  breakable
]
Instead of generating pseudo-target domains during training, we propose using generative models for image editing (e.g., Flux.1 Kontext~\cite{labs2025flux1kontextflowmatching} or Qwen-Image-Edit~\cite{wu2025qwenimagetechnicalreport}) to transform target domain images back to the source domain at test time:
\begin{equation}
    x_n^{PS}= G(x_n^T,t^S)
    \label{eq::xps}
\end{equation}
where $t^S$ describes the source domain distribution.
\textbf{TTM assumes no access to target-domain data beyond the current test image and performs no model adaptation.}
\end{tcolorbox}

This approach offers a key advantage: rather than requiring $K$ prompts $t_{1\dots K}$ to describe all possible unknown target domains, we need only \textit{a single prompt} $t^S$ describing the known source domain. This dramatically reduces complexity from describing multiple unknown distributions to characterizing one well-understood distribution.

At test time, we apply the discriminative model $f_\theta$ to the pseudo-source-domain image:
\begin{equation}
    \mathbf{y}_n^{PS}=f_\theta(x_n^{PS})=f_\theta(G(x_n^T,t^S))
\end{equation}

Since $f_\theta$ was trained on the source domain distribution, predictions on $x_n^{PS}$ should be more accurate than those obtained directly from $x_n^T$. Crucially, this approach requires no additional training 
of the discriminative model.

One potential challenge is maintaining semantic consistency during the inverse transformation. The generative model $G$ may inadvertently introduce or remove semantic content—a known limitation affecting even source-to-target transformation methods. To address this, we ensemble predictions from both the original and pseudo-source domain images:
\begin{equation}
        \mathbf{y}_n^{T}=0.5f_\theta(x_n^{PS}) +0.5 f_\theta(x_n^{T})
    \label{eq::fusion}
\end{equation}
This fusion strategy grounds the final prediction in the original image while benefiting from the domain-aligned features of the transformed image. Note that this fusion step is task-specific—we apply it for semantic segmentation but not for object detection or image classification.

\begin{figure*}[t]
  \centering
  \includegraphics[width=1.0\textwidth]{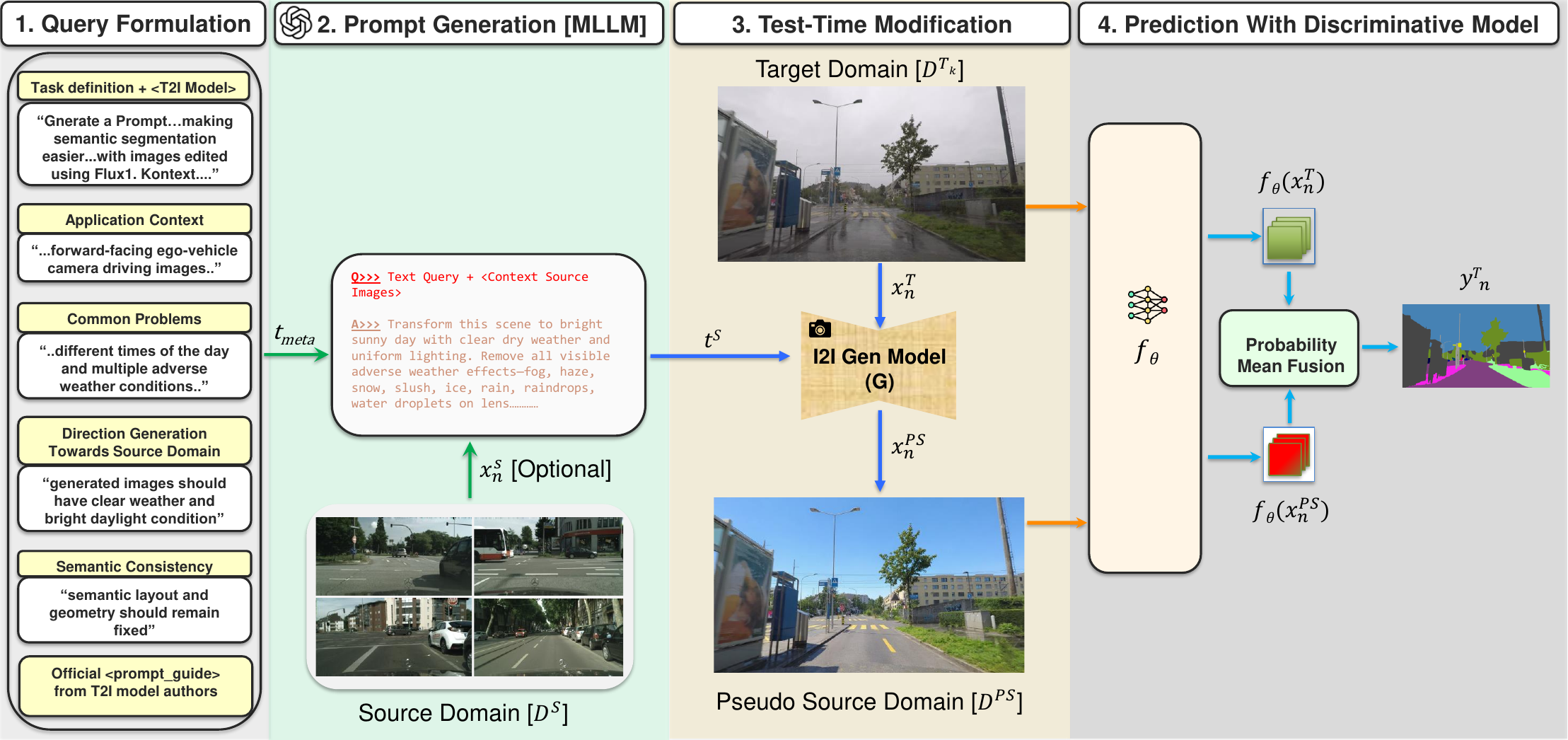}
  \caption{Overview of our test-time modification pipeline doing inverse domain transfer. The process begins with formulating a meta-prompt that describes the desired transformation. An MLLM processes this meta-prompt to generate an effective prompt $t^S$ for the I2I generation model. During test time, the I2I model transforms target domain images to the source domain. Finally, the pre-trained discriminative model processes both images, with task-specific fusion strategies applied as needed.}
  \label{fig:piepline_figure}
  \vspace{-1.5em}
\end{figure*}

\noindent \textbf{Generating Effective Source Domain Prompts.}
\label{para:query_formulation}
To obtain an effective source domain prompt $t^S$, we employ the two-stage pipeline illustrated in~\cref{fig:piepline_figure}.

\textit{Step 1: Meta-Prompt Formulation.} We construct a comprehensive meta-prompt: a human-written text, denoted as $t_\text{meta}$, designed to generate another prompt. This meta-prompt instructs a multimodal large language model (MLLM), such as GPT-5~\cite{openai_gpt5}, to generate a prompt suitable for inverse domain transformation using an Image-to-Image (I2I) generation model.
The meta-prompt $t_\text{meta}$ integrates the following components:
\begin{itemize}[topsep=2pt,itemsep=1pt]
    \item \textbf{Task specification}: The downstream task (e.g., semantic segmentation)
    \item \textbf{Model information}: The specific I2I model being used
    \item \textbf{Objective}: The intended goal (e.g., improving segmentation on test images)
    \item \textbf{Domain context}: Application-specific information (\eg ``road driving scenes'')
    \item \textbf{Expected challenges}: Common domain shifts such as weather conditions or lighting variations
    \item \textbf{Transformation requirements}: Instructions to preserve semantic layout while adapting appearance
    \item \textbf{Model guidelines}: Best practices from the I2I model's documentation
\end{itemize}

While specifying expected challenges requires minimal prior knowledge, this information could alternatively be inferred at test time using MLLMs. Optionally, we can provide a small subset of $m$ source domain images $\{x_1^{S},\dots, x_m^{S}\}$ as visual context, though this is typically unnecessary for well-known datasets.

\textit{Step 2: Prompt Generation.} The MLLM processes the meta-prompt to generate the final source domain prompt:
\begin{equation}
    t^{S} = \mathrm{MLLM}(t_\text{meta}, \{x_1^{S},\dots, x_m^{S}\})
\end{equation}

The resulting $t^{S}$ enables transformation of target domain images into semantically consistent pseudo-source domain images. For example, in our driving scene experiments, we use the following prompt that removes various domain shift artifacts while preserving scene structure:

\noindent\fbox{\parbox{0.97\columnwidth}{\small\texttt{Transform this scene to bright sunny day with clear dry weather and uniform lighting. Remove all visible adverse weather effects [...] while maintaining original object positions, scale, scene composition, camera angle, and framing exactly as in the input.}}}

\noindent We provide complete prompt variations and ablation studies in the appendix.

\vspace{-0.5em}   

\subsection{Intuition: Reducing Aleatoric Uncertainty}
\label{sec:aleatoric}

Our approach not only addresses epistemic uncertainty (model uncertainty) but also reduces aleatoric uncertainty inherent in the data. For a discriminative model $f_\theta(x)$, the predictive variance decomposes as:
\begin{equation}
    \mathrm{Var}[y \mid x] = \underbrace{\mathrm{Var}_{\theta}[f_\theta(x)]}_{\text{epistemic}} + \underbrace{\mathbb{E}_{\theta}[\mathrm{Var}(y \mid x, \theta)]}_{\text{aleatoric}}
\end{equation}

While epistemic uncertainty can be reduced through additional training data, aleatoric uncertainty arises from the input itself—due to occlusions, low light, noise, or adverse weather—and cannot be reduced by collecting more training samples:
\begin{equation}
    \mathrm{Var}(y \mid x, \theta) = \mathrm{Var}(y \mid x, \theta^+)
\end{equation}
where $\theta^+$ represents parameters learned from additional training data.

\textbf{Inverse Transformation as Aleatoric Reduction.}
Modern generative models, trained on diverse image corpora, possess broad prior knowledge similar to human visual experience. By transforming a corrupted target image $x^T$ to a cleaner pseudo-source image $x^{PS}$, we effectively reduce the input-dependent uncertainty:
\begin{equation}
    \mathrm{Var}(y \mid x^{PS}, \theta) < \mathrm{Var}(y \mid x^T, \theta)
\end{equation}

This allows our test-time modification to mitigate effects such as snow cover, raindrops, or low-light conditions (see~\cref{fig:qual_results_segm} and~\cref{fig:qual_results_det}). Importantly, this reduction in aleatoric uncertainty would not be achievable by synthesizing pseudo-target domains during training, as aleatoric uncertainty is fundamentally tied to the input rather than the model.

\section{Experiments}
\label{sec:experiments}
\begin{table*}[t]
\centering
\caption{Semantic segmentation performance (mIoU in \%) on different domain generalization benchmarks with and without (base model) using test-time modification (TTM). CS refers to Cityscapes and Geo. shift refers to geographic shift.}
\vspace{-0.5em}
\label{tab:A_main_table_summary}
\resizebox{\textwidth}{!}{
\begin{tabular}{l|ccccc|c>{\centering\arraybackslash}m{1.3cm}}
\hline
\textbf{Method} & \makecell{\textbf{DeepLabV3+} \\ \textbf{[ResNet-101-d8]}} & \makecell{\textbf{OCRNet} \\ \textbf{[HRNetV2-W48]}} & \makecell{\textbf{Segformer} \\ \textbf{[MiT-B1]}} & \makecell{\textbf{Segformer} \\ \textbf{[MiT-B5]}} & \makecell{\textbf{Mask2Former} \\ \textbf{[Swin-L]}} & \textbf{mIoU$^\text{Avg}$} & \textbf{$\Delta$$\uparrow$} \\
\hline
\addlinespace[2pt]
\multicolumn{8}{c}{\textbf{Clear-to-Adverse-Weather:} \textbf{CS $\rightarrow$ ACDC}} \\
\hline
\addlinespace[2pt]
Base Model & 42.0 & 45.5 & 46.9 & 57.1 & 60.5 &  50.4 &  \\
+ Flux1. Kontext Max (TTM) & 56.5 & 56.0 & 55.7 & 61.3 & 64.3 & 58.8 &  \\
+ QIE-2509 (TTM) & \underline{58.3} & \underline{58.9} & \underline{58.9} & \underline{64.6} & \underline{66.4} & \textbf{61.4} & \textcolor{green!60!black}{\textbf{+11.0}} \\
\hline
\addlinespace[2pt]
\multicolumn{8}{c}{\textbf{Day-to-Nighttime: CS $\rightarrow$ DarkZurich}} \\
\addlinespace[2pt]
\hline
Base Model & 19.8 & 22.9 & 23.2 & 36.8 & 40.6 & 28.6 &  \\
+ Flux1. Kontext Max (TTM) & 41.2 & 40.9 & 43.1 & \underline{47.9} & \underline{51.7} & 45.0 &  \\
+ QIE-2509 (TTM) & \underline{44.8} & \underline{43.9} & \underline{44.9} & 46.7 & 51.3 & \textbf{46.3} & \textcolor{green!60!black}{\textbf{+17.7}} \\
\hline
\addlinespace[2pt]
\multicolumn{8}{c}{\textbf{Day-to-Nighttime + Geo Shift: CS $\rightarrow$ BDD100K-Night}} \\
\addlinespace[2pt]
\hline
Base Model & 23.8 & 24.4 & 25.8 & 34.0 & 40.6 & 29.7 &  \\
+ Flux1. Kontext Max (TTM) & 42.3 & 37.6 & \underline{41.9} & 44.7 & \underline{49.1} & 43.1 &  \\
+ QIE-2509 (TTM) & \underline{42.8} & \underline{43.0} & 41.8 & \underline{46.6} & 47.1 & \textbf{44.3} & \textcolor{green!60!black}{\textbf{+14.6}} \\
\hline
\addlinespace[2pt]
\multicolumn{8}{c}{\textbf{Clear-to-Adverse-Weather + Geo Shift: CS $\rightarrow$ BDD100K}} \\
\addlinespace[2pt]
\hline
Base Model & 48.8 & 49.7 & 47.4 & 55.2 & 58.0 & 51.8 &  \\
+ Flux1. Kontext Max (TTM) & \underline{53.2} & \underline{52.8} & \underline{50.5} & \underline{55.8} & \underline{58.7} & \textbf{54.2} & \textcolor{green!60!black}{\textbf{+2.4}} \\
+ QIE-2509 (TTM) & 50.7 & 50.5 & 48.7 & 53.7 & 55.0 & 51.7 &  \\
\hline
\end{tabular}
}
\vspace{-0.5em}
\end{table*}

\paragraph{Datasets}
Across all downstream tasks, we use a total of six target datasets for evaluation and two datasets for training our source domain models. For semantic segmentation, we use the clear weather Cityscapes~\cite{cordts2016cityscapes} as our source dataset. Whereas we evaluate on adverse weather and nighttime datasets - ACDC~\cite{sakaridis2021acdc}, BDD100K~\cite{yu2020bdd100k}, BDD100K-Night~\cite{sakaridis2020map, yu2020bdd100k}, and Dark Zurich~\cite{sakaridis2019guided}. For the object detection task, we again use Cityscapes to train our source model and evaluate on a randomly sampled subset of 501 BDD100K nighttime images with object detection annotations. We refer to this validation set as BDD100K-Night-Det. For image classification,  we use the ImageNet-1K~\cite{deng2009imagenet} dataset as our source domain and ImageNet-R~\cite{hendrycks2021many} dataset as our target set. The latter contains difficult, rendered image versions of a subset of 200 categories. Refer to the appendix for more information on the various datasets.

\paragraph{Implementation Details}
For our semantic segmentation experiments, we utilize both convolutional neural networks (CNNs) and transformer-based models of varying sizes. CNN based models include DeepLabV3+~\cite{chen2018encoder} and OCRNet~\cite{yuan2020object} whereas transformer based models include Segformer~\cite{xie2021segformer} and Mask2Former~\cite{cheng2022masked}. All these models are pre-trained on the Cityscapes dataset and are directly evaluated on the validation set of the target dataset. For object detection evaluation, we use the Faster-RCNN~\cite{ren2015faster} and Mask-RCNN~\cite{he2017mask} models. For the image classification task, we use the ResNet-50 and ResNet-152 models~\cite {he2016deep}. For image generation, we mainly utilize two different image-to-image (I2I) generation models as part of TTM, including a commercial model, Flux.1 Kontext Max and an open-source model Qwen-Image-Edit-2509 [QIE-2509]. We also do an ablation study on additional I2I generation models including Flux.2 variants~\cite{blackforestlabs2025flux2}. More details on different models are provided in the appendix.

\subsection{Semantic Segmentation}

\begin{figure*}[t]
  \centering
  \scriptsize
\setlength\tabcolsep{3.8pt}  %
\begin{tabularx}{\linewidth}{p{0.1785\linewidth}p{0.1785\linewidth}p{0.1785\linewidth}p{0.1785\linewidth}p{0.1785\linewidth}}
\centering Original Image & 
\centering w/o TTM& 
\centering TTM Image & 
\centering w/ TTM & 
\centering\arraybackslash Ground Truth \\
\end{tabularx}

  \includegraphics[width=1.0\linewidth]{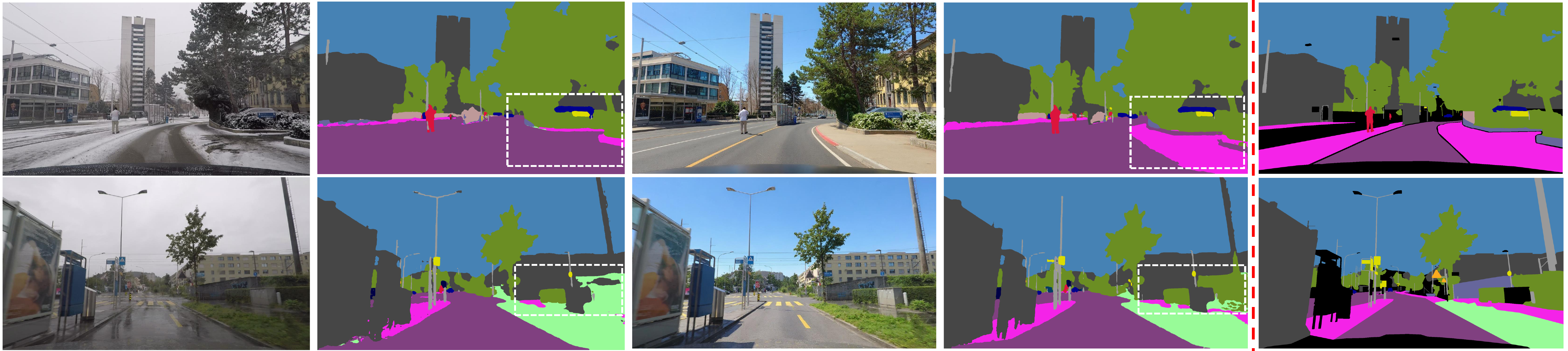}
    \scriptsize%
\setlength\tabcolsep{1pt}%
\resizebox{\linewidth}{!}{%
\newcolumntype{P}[1]{>{\centering\arraybackslash}p{#1}}
\begin{tabular}{@{}*{20}{P{0.065\linewidth}}@{}}
     {\cellcolor[rgb]{0.5,0.25,0.5}}\textcolor{white}{road} 
     &{\cellcolor[rgb]{0.957,0.137,0.91}}sidew. 
     &{\cellcolor[rgb]{0.275,0.275,0.275}}\textcolor{white}{build.} 
     &{\cellcolor[rgb]{0.4,0.4,0.612}}\textcolor{white}{wall} 
     &{\cellcolor[rgb]{0.745,0.6,0.6}}fence 
     &{\cellcolor[rgb]{0.6,0.6,0.6}}pole 
     &{\cellcolor[rgb]{0.98,0.667,0.118}}t.light
     &{\cellcolor[rgb]{0.863,0.863,0}}t.sign 
     &{\cellcolor[rgb]{0.42,0.557,0.137}}veget. 
     &{\cellcolor[rgb]{0.596,0.984,0.596}}terrain 
     &{\cellcolor[rgb]{0.275,0.510,0.706}}sky
     &{\cellcolor[rgb]{0.863,0.078,0.235}}\textcolor{white}{person} 
     &{\cellcolor[rgb]{0.988,0.494,0.635}}\textcolor{black}{rider} 
     &{\cellcolor[rgb]{0,0,0.557}}\textcolor{white}{car} 
     &{\cellcolor[rgb]{0,0,0.275}}\textcolor{white}{truck} 
     &{\cellcolor[rgb]{0,0.235,0.392}}\textcolor{white}{bus}
     &{\cellcolor[rgb]{0,0.392,0.471}}\textcolor{white}{train} 
     &{\cellcolor[rgb]{0,0,0.902}}\textcolor{white}{m.bike} 
     & {\cellcolor[rgb]{0.467,0.043,0.125}}\textcolor{white}{bike}
     &{\cellcolor[rgb]{0,0,0}}\textcolor{white}{n/a.}
\end{tabular}
}

  \caption{Qualitative performance comparison for domain generalized semantic segmentation on ACDC dataset - with and without using TTM. The same, unmodified Mask2Former trained on Cityscapes is used as the segmentation model for inference. Dashed boxes show how TTM can reduce aleatoric uncertainty of the modified image by incorporating world knowledge, leading to improved predictions.}
  \label{fig:qual_results_segm}
  \vspace{-0.5em}
\end{figure*}

\begin{figure*}[t]
  \centering

  \begin{subfigure}{0.28\textwidth}
    \centering
    \includegraphics[width=\linewidth, trim=0 0 66.7 0, clip]{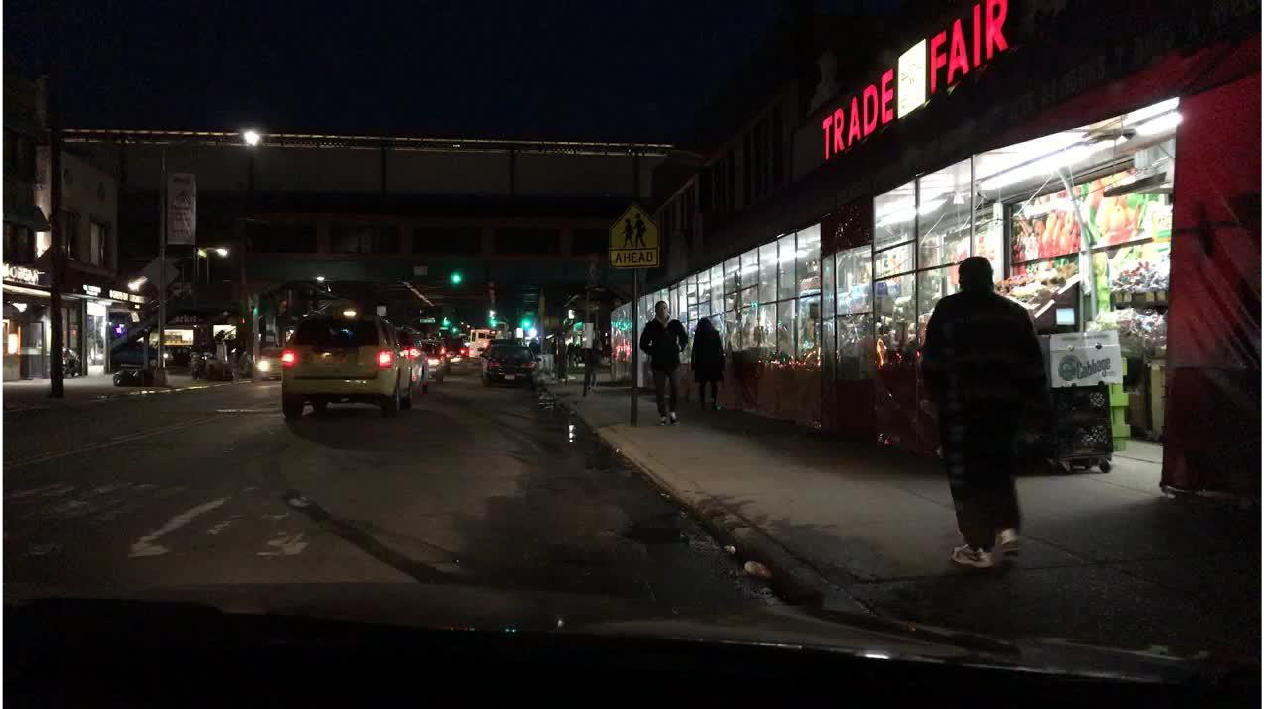}
    \caption{Original Image}
    \label{fig:row1_left}
  \end{subfigure}\hfill
  \begin{subfigure}{0.28\textwidth}
    \centering
    \includegraphics[width=\linewidth, trim=33.3 0 33.3 0, clip]{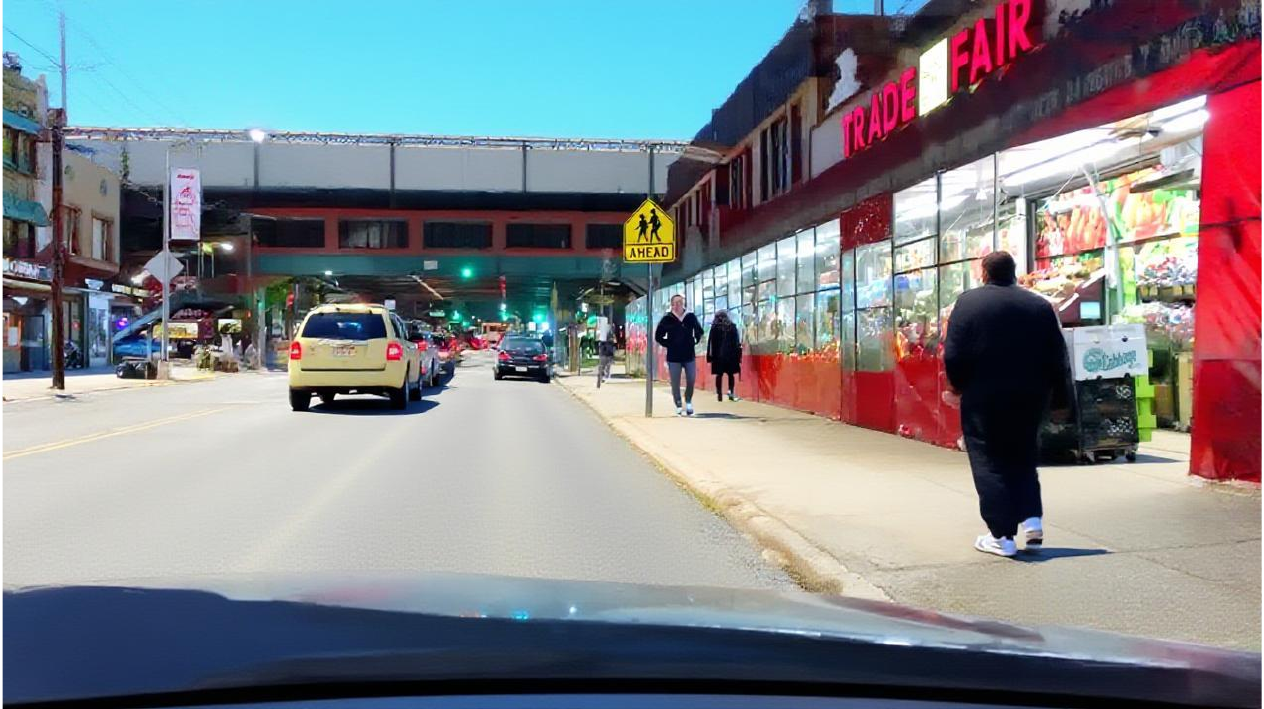}
    \caption{Flux--1 Kontext Dev}
    \label{fig:row1_mid}
  \end{subfigure}\hfill
  \begin{subfigure}{0.28\textwidth}
    \centering
    \includegraphics[width=\linewidth, trim=66.7 0 0 0, clip]{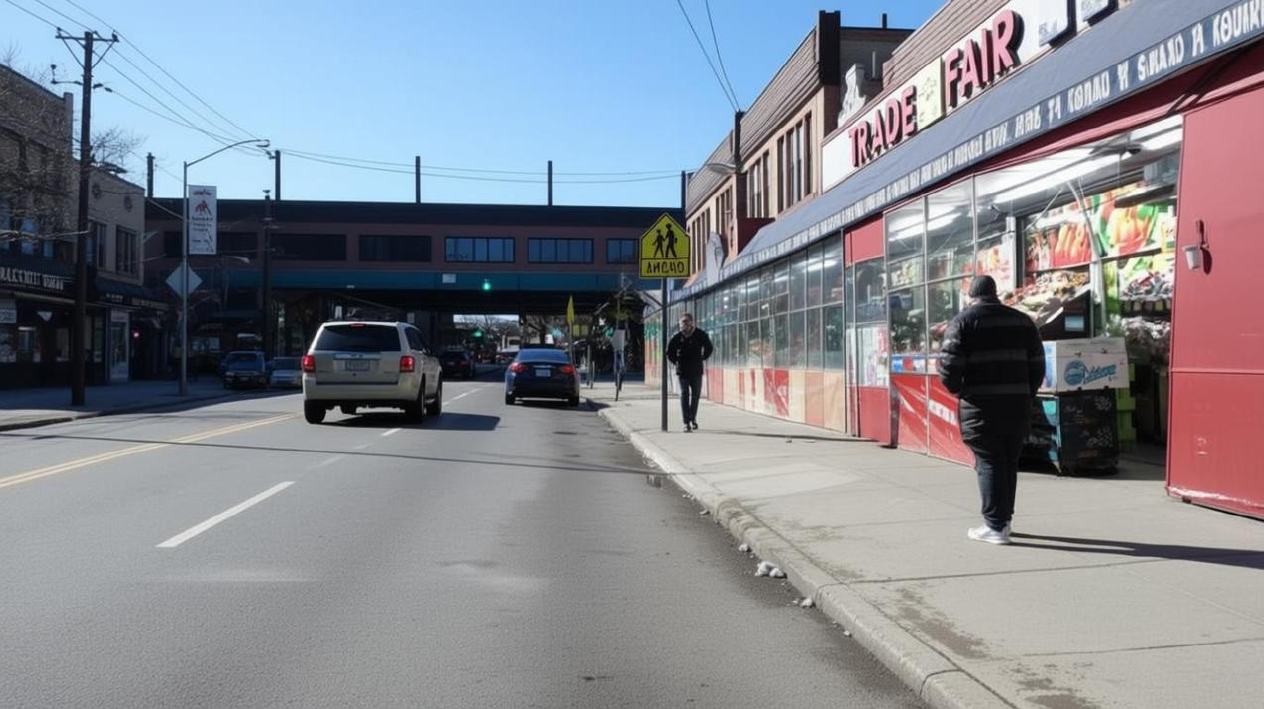}
    \caption{QIE-2509}
    \label{fig:row1_right}
  \end{subfigure}

  \begin{subfigure}{0.28\textwidth}
    \centering
    \includegraphics[width=\linewidth, trim=0 0 66.7 0, clip]{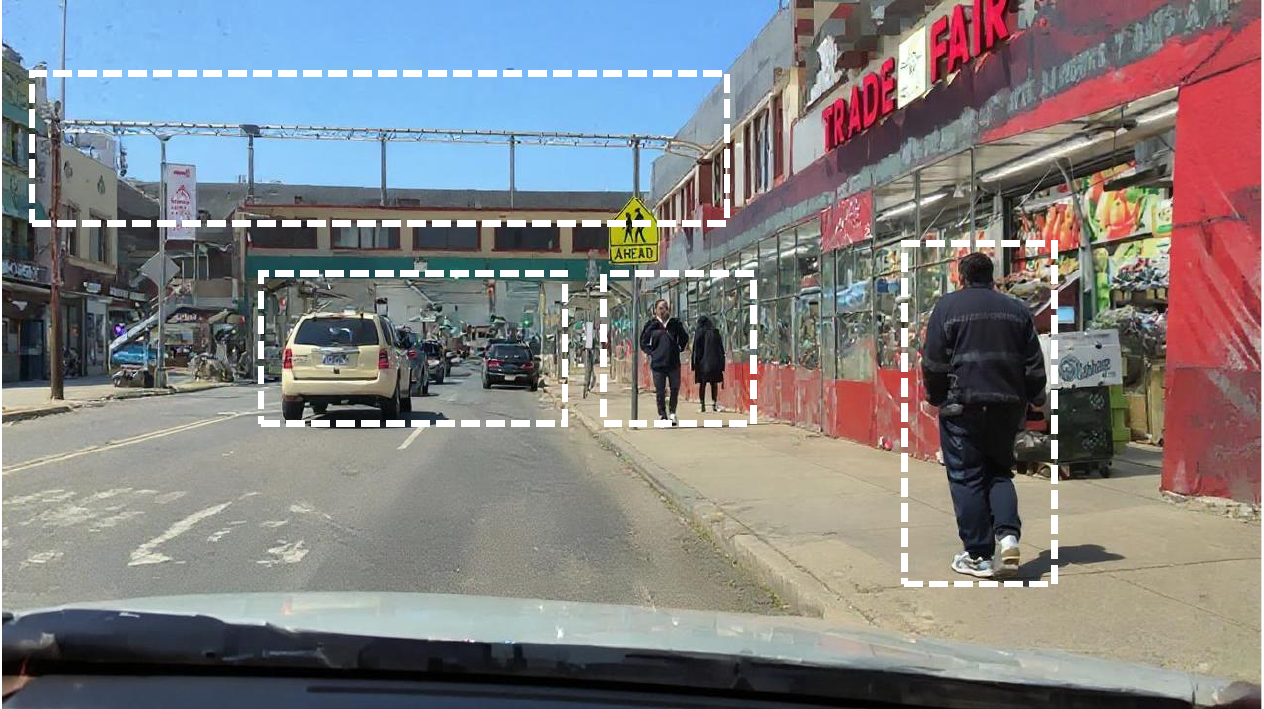}
    \caption{Flux.1 Kontext Max}
    \label{fig:row2_left}
  \end{subfigure}\hfill
  \begin{subfigure}{0.28\textwidth}
    \centering
    \includegraphics[width=\linewidth, trim=33.3 0 33.3 0, clip]{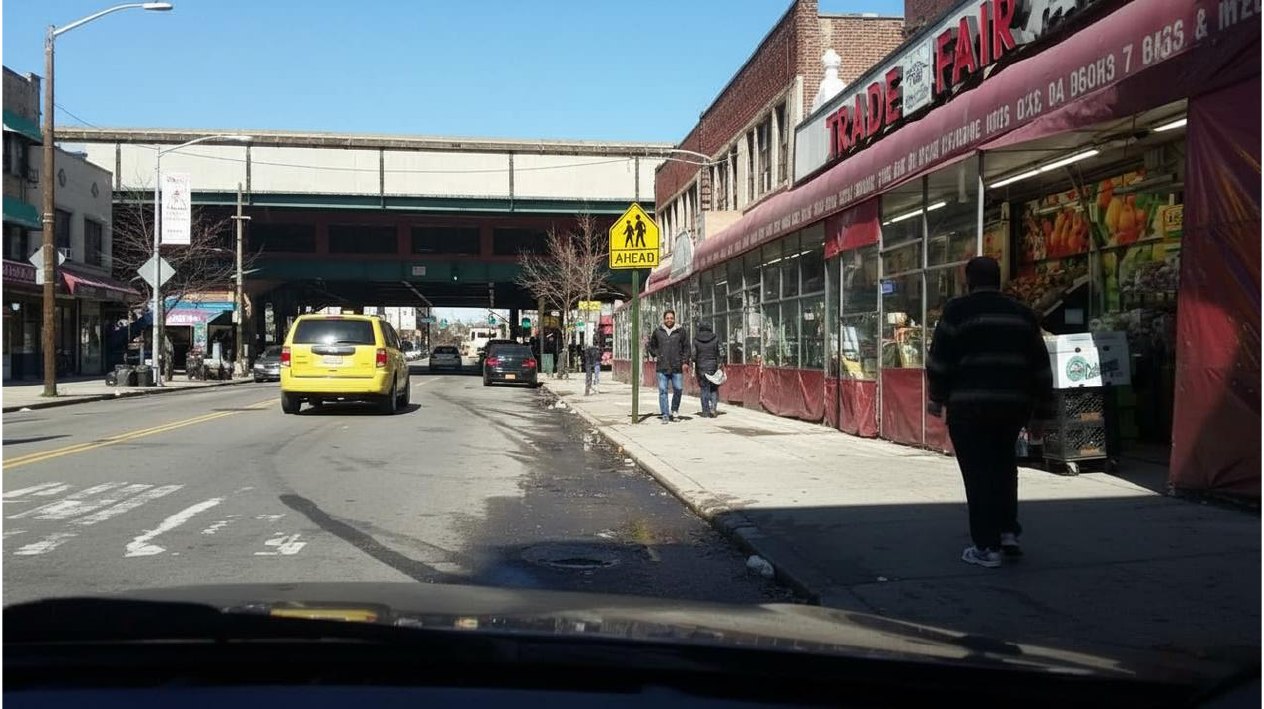}
    \caption{Nano Banana}
    \label{fig:row2_mid}
  \end{subfigure}\hfill
  \begin{subfigure}{0.28\textwidth}
    \centering
    \includegraphics[width=\linewidth, trim=66.7 0 0 0, clip]{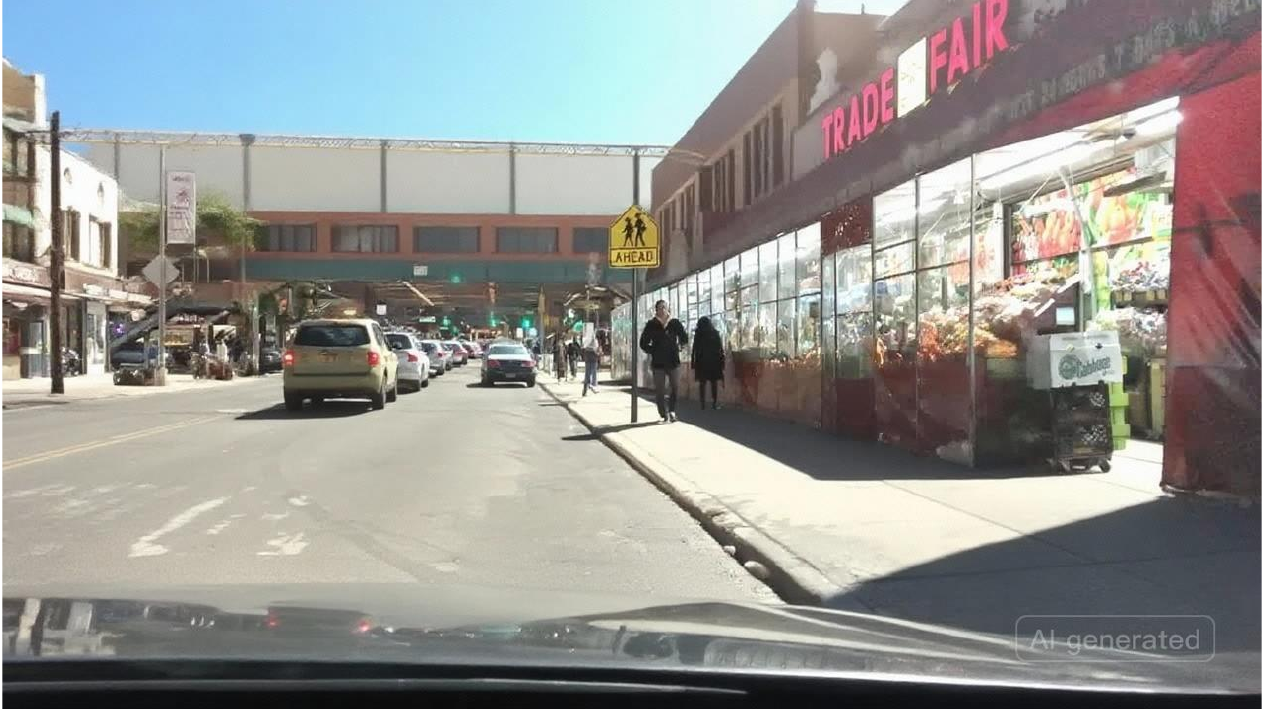}
    \caption{SeedEdit 3.0}
    \label{fig:row2_right}
  \end{subfigure}

  \caption{Qualitative comparison of the generation quality of test-modified images among different I2I generation models on BDD100K-Night. Dashed boxes highlight key layout components that are not respected by some generations.}

  \label{fig:gen_mod_comp}
  \vspace{-1.5em}
\end{figure*}

\begin{figure}[t]
  \centering
  \begin{subfigure}[t]{0.245\columnwidth}  %
    \centering
    \includegraphics[width=\linewidth]{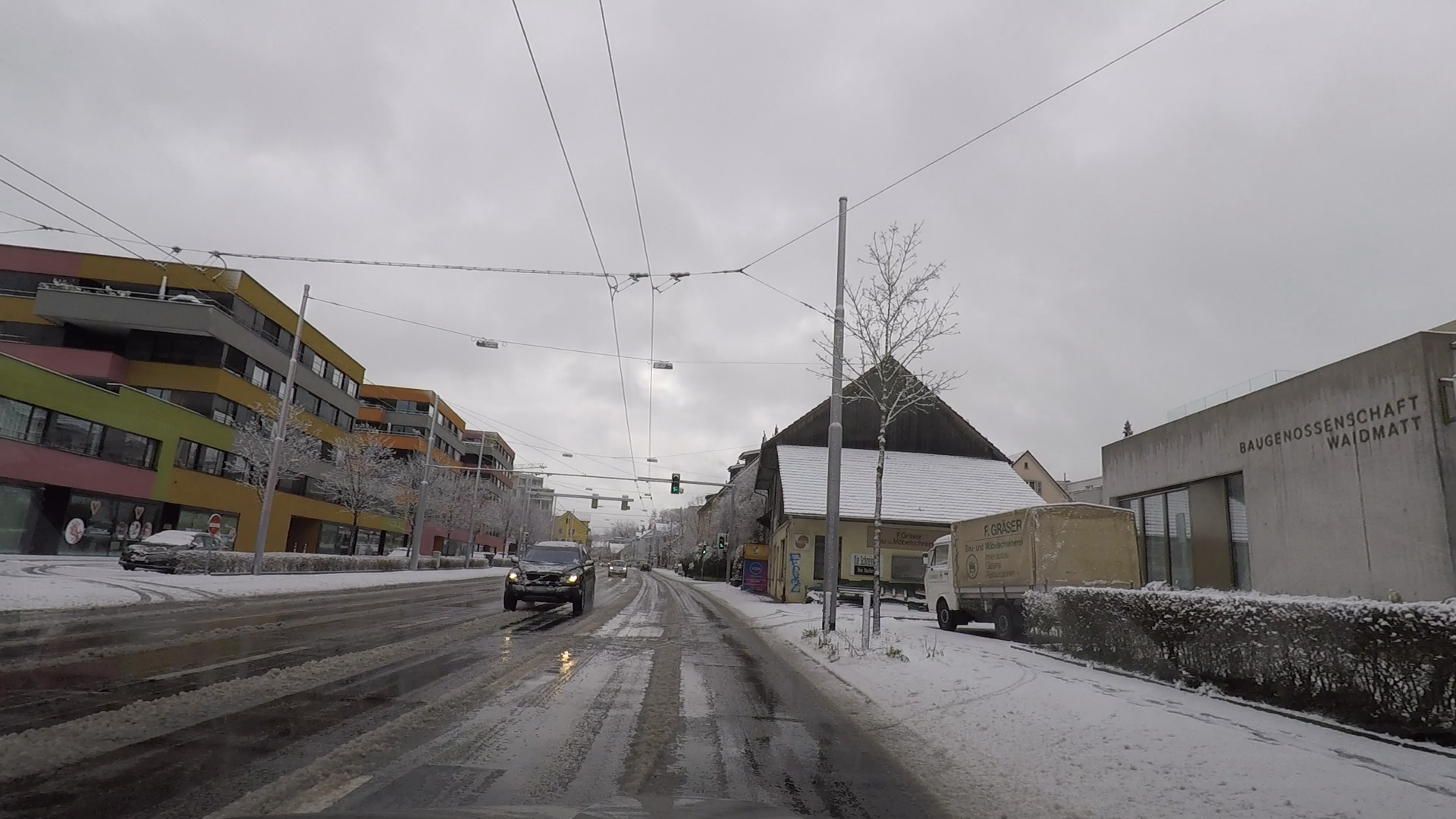}
    \caption{Original Image}
    \label{fig:r1_a}
  \end{subfigure}\hfill
  \begin{subfigure}[t]{0.245\columnwidth}
    \centering
    \includegraphics[width=\linewidth]{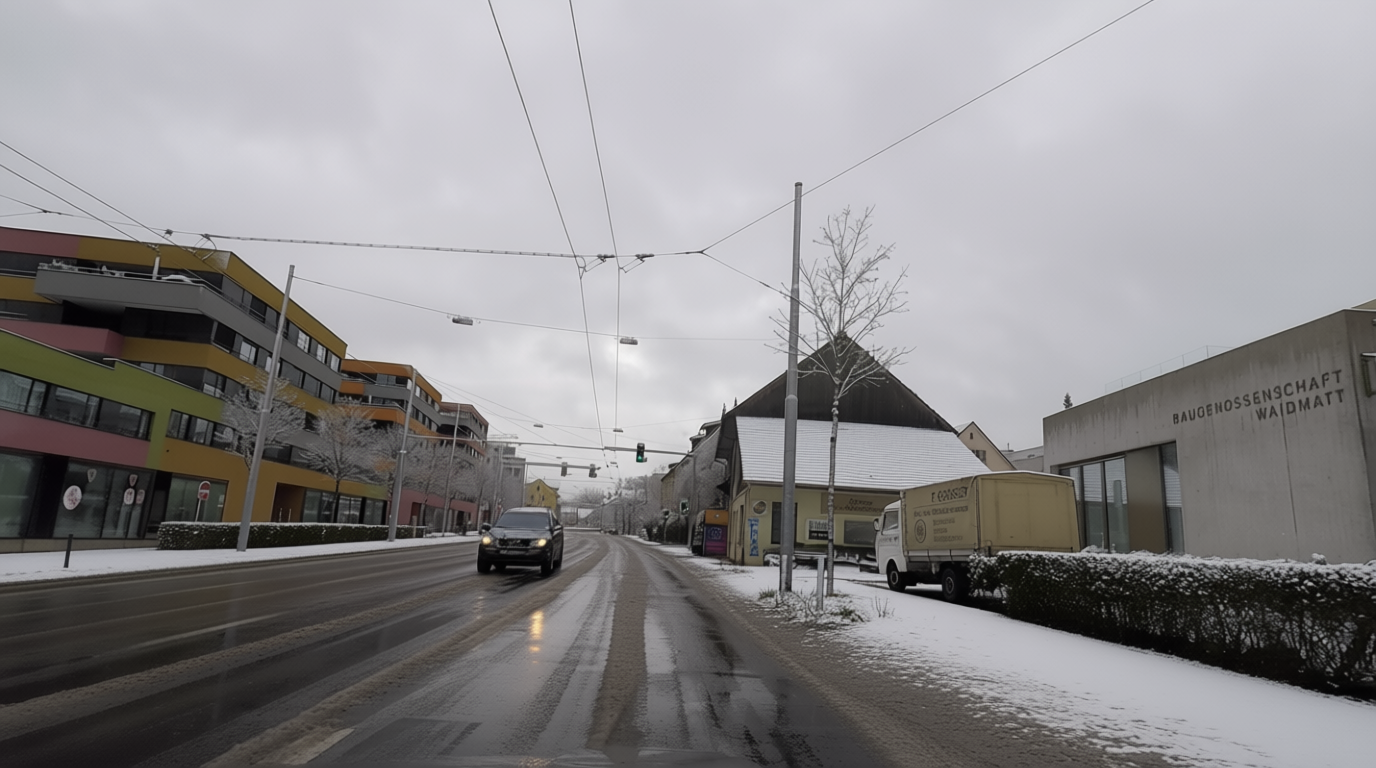}
    \caption{Handcrafted}
    \label{fig:r1_b}
  \end{subfigure}\hfill
  \begin{subfigure}[t]{0.245\columnwidth}
    \centering
    \includegraphics[width=\linewidth]{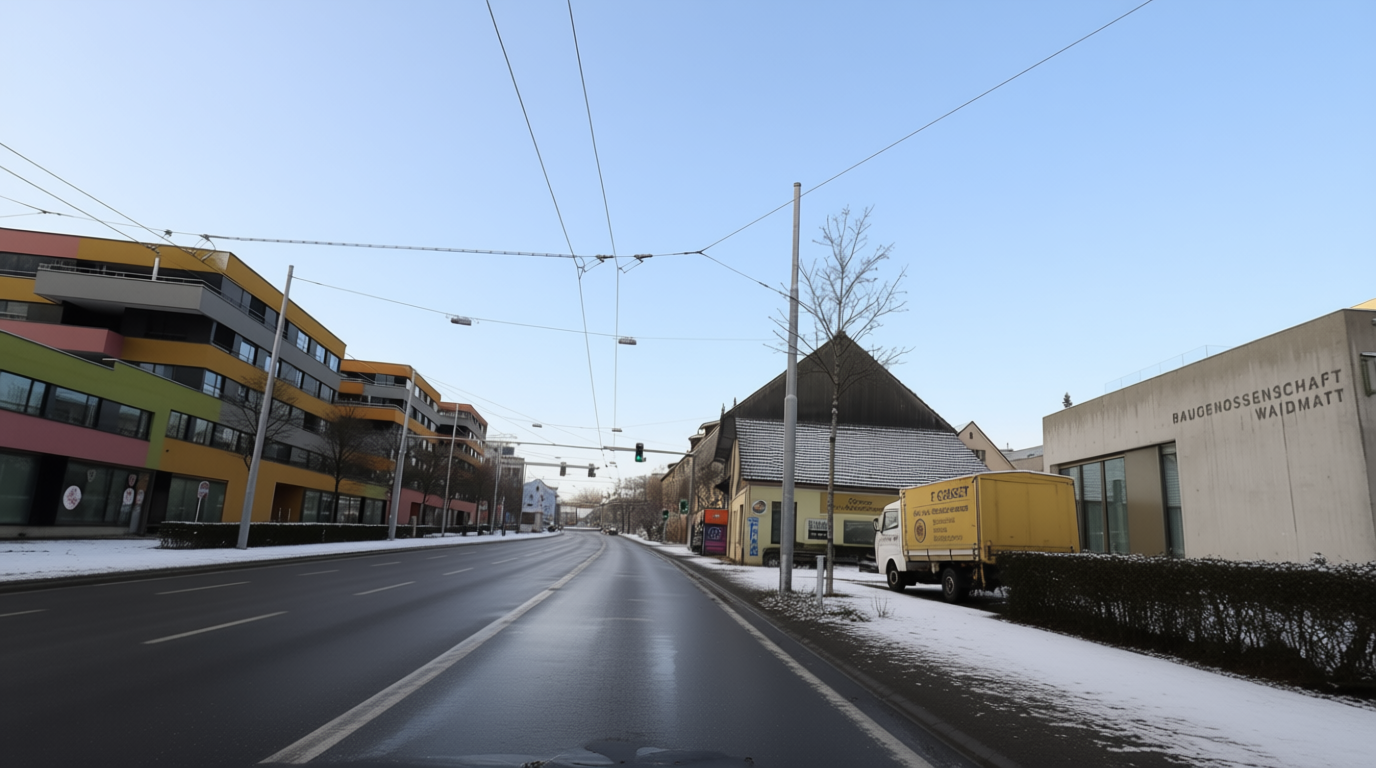}
    \caption{Metaprompt LLM}
    \label{fig:r1_c}
  \end{subfigure}\hfill
  \begin{subfigure}[t]{0.245\columnwidth}
    \centering
    \includegraphics[width=\linewidth]{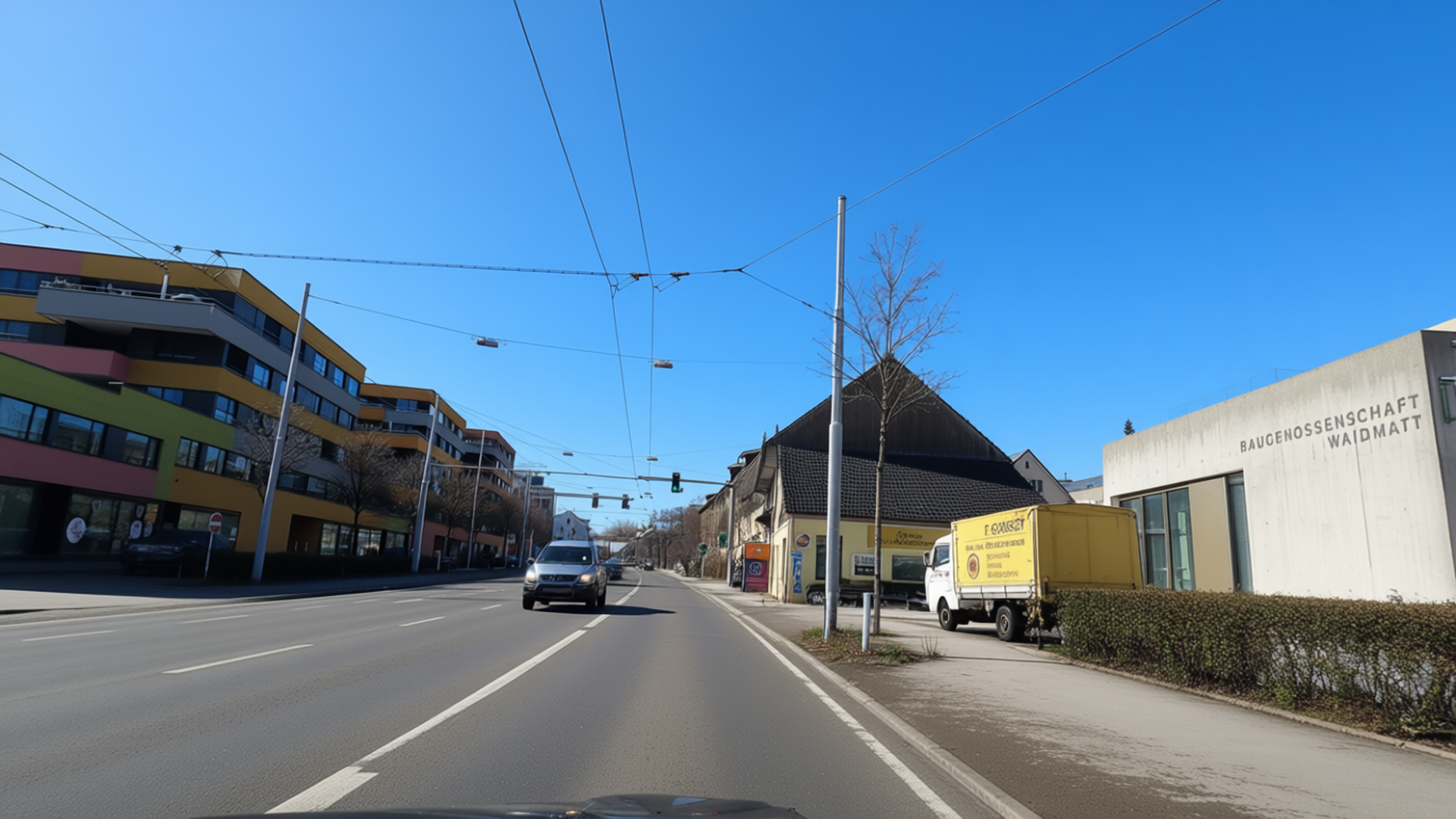}
    \caption{Ours (TTM)}

    \label{fig:r1_d}
  \end{subfigure}
  \caption{Comparing prompting strategies for inverse domain transfer using QIE-2509. Handcrafted prompts (a) often fail to produce the desired inverse transformation. Metaprompting an MLLM like GPT-5 (c) comes closer to achieving the desired inverse transformation but still deviates while often compromising the semantic consistency, \eg missing a car on the road. Metaprompting an MLLM with our TTM formulation (d) carries concrete guidance necessary to achieve the desired transformation without trading off the semantic consistency.}
  \label{fig:prompting_ablation}
  \vspace{-0.5em}
\end{figure}

\begin{figure}[t]
  \centering
  \scriptsize
\setlength\tabcolsep{4.5pt}  %
\begin{tabularx}{\linewidth}
  {p{0.31\linewidth}p{0.31\linewidth}p{0.31\linewidth}}  %
  \centering Image w/ GT Labels & 
  \centering w/o TTM & 
  \centering\arraybackslash w/ TTM \\
\end{tabularx}

  \includegraphics[width=0.98\columnwidth]{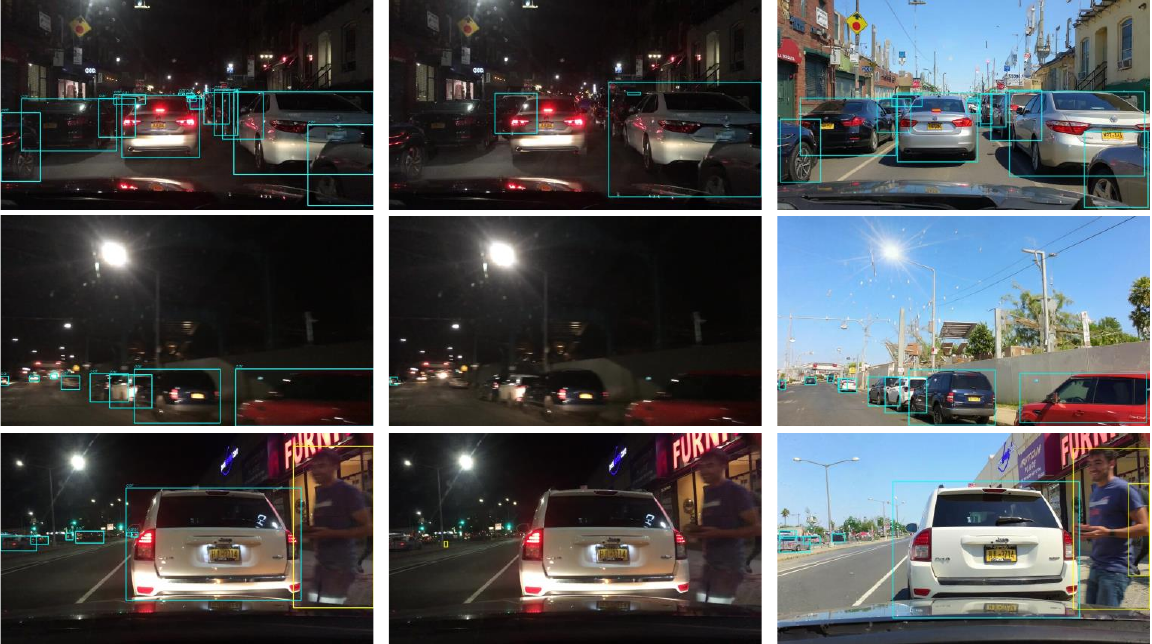}
  \caption{Qualitative performance comparison for domain generalized object detection on Cityscapes $\rightarrow$ BDD100KNight-Det using Faster-RCNN - without and with using TTM. Three different examples from BDD100KNight-Det dataset.}
  \label{fig:qual_results_det}
  \vspace{-2.2em}
\end{figure}

\begin{table}[t]
  \centering
  {\scriptsize
  \setlength{\tabcolsep}{3pt}  %
  \caption{CS → DarkZurich: mIoU across 10 seeds (mean ± standard deviation) for 10 different versions of DarkZurich generated using QIE-2509 with the same text prompt. The second row reports average mIoU with standard deviation.}
  \label{tab:D_variance_10_seed_gen}
  \vspace{-0.5em}
  \begin{tabular}{@{}l *{5}{c}@{}}
    \toprule
    & DeepLabV3+ & OCRNet & MiT-B1 & MiT-B5 & Mask2Former \\
    \midrule
    \textbf{w/o TTM}     & 19.8 & 22.9 & 23.2 & 36.8 & 40.6 \\
    \textbf{w/ TTM}      & 43.7 $\pm$ 1.3 & 43.1 $\pm$ 1.3 & 42.9 $\pm$ 1.6 & 47.5 $\pm$ 1.3 & \textbf{50.5 $\pm$ 1.2} \\
    \bottomrule
  \end{tabular}
  }
  \vspace{-0.5em}
\end{table}

We evaluate on the domain generalized semantic segmentation task in the autonomous driving context, focusing on weather and daytime shifts, as they cause the most significant domain gap in autonomous driving applications. \Cref{tab:A_main_table_summary} shows a performance comparison of our Test-Time Modification (TTM) method on the domain generalized semantic segmentation task. As shown in the table, using TTM yields significant improvements in mIoU across various datasets and segmentation models. This performance jump is even more significant with smaller models, such as DeepLabV3+, OCRNet, and Segformer [MiT-B1]. This demonstrates that weaker models can be made more robust by using TTM. An interesting observation that supports this reasoning is that the weakest DeepLabV3+ model, when used with TTM, outperforms a much stronger Segformer [MiT-B5] model for the CS $\rightarrow$ ACDC benchmark. The same trend persists with CS $\rightarrow$ DarkZurich and CS $\rightarrow$ BDD100K-Night. Also, for CS $\rightarrow$ BDD100K, Flux1. Kontext Max outperforms QIE-2509 likely because it is more robust to geographical shifts, which is the primary change in BDD100K w.r.t Cityscapes. Qualitative results for domain generalized semantic segmentation on different target datasets are shown in~\cref{fig:qual_results_segm}.

\vspace{-1.0em}

\subsubsection{Ablations.} We perform an ablation study with QIE-2509 in~\cref{tab:D_variance_10_seed_gen}, where we generate 10 different Dark Zurich validation sets (for 10 random seeds with the same prompt) using TTM and report the average mIoU across these generations with standard deviation. We try to study the scale of randomization in these I2I (Image-to-Image) generation models. Considering a small validation set (50 images), the deviation is reasonable, and we are still able to make significant improvements. We perform another ablation in~\cref{tab:E_diff_gen_models_mIoU} where we use five different I2I generation models and compare the segmentation performance on the BDD100K-Night dataset. We use the same prompt for all generative models, obtained during the TTM process. Flux. 1 Kontext Max performs the best when used for TTM. We show a comparison of the generation quality of test-modified images between different generation models in~\cref{fig:gen_mod_comp}. As can be seen, all models are capable of performing night-to-day transfer with good fidelity. However, Flux.1 Kontext Max maintains the best semantic consistency among all models, as highlighted in the figure. In another ablation study, we show the effect of correct prompting in~\cref{fig:prompting_ablation} by generating images using QIE-2509. Handcrafted short prompts, such as "remove snow from the image," often do not yield the desired output. Metaprompting an MLLM, such as GPT-5, is a better strategy; however, without a precise query, an MLLM can generate weak prompts. As we see in~\cref{fig:r1_c}, the car on the road also gets removed, resulting in a significant layout mismatch. In contrast, our query formulation from the TTM process, as described in~\cref {para:query_formulation}, results in complete snow removal while preserving the semantic layout. We also performed an ablation to show the effectiveness of the probability mean fusion step described in~\cref{sec:method}. \Cref{tab:g_fusion_ablation} shows the ablation results for semantic segmentation. Across all evaluated domain shifts, fusion consistently improves TTM performance. Segmentation allows simple pixel-level fusion; whereas detection fusion requires box matching to avoid duplicates and adds task-specific complexity. Conversely, image classification lacks the detailed spatial constraints that necessitate fusion step in the first place. Thus, we only apply the fusion step for semantic segmentation task.

\vspace{-1.1em}

\begin{table}[h]
  \centering
  {\scriptsize  %
  \setlength{\tabcolsep}{4pt}  %
  \renewcommand{\arraystretch}{1.3}  %
  \caption{CS → BDD100K-Night: segmentation performance comparison for different I2I generative models with Mask2Former segmentation model.
  Abbreviations: F1.K.Dev=FLUX.1 Kontext [Dev]; NB=Nano-Banana; SE3=SeedEdit 3.0; QIE=Qwen-Image-Edit-2509; F2=FLUX.2 [Dev]; F1.K.Max=FLUX1. Kontext [Max]. }
  \label{tab:E_diff_gen_models_mIoU}
  \vspace{-0.5em}
  \begin{tabular}{@{}lccccccc@{}}
    \toprule
    & w/o TTM & F1.K.Dev~\cite{labs2025flux1kontextflowmatching} & NB~\cite{comanici2025gemini} & SE3~\cite{wang2025seededit} & QIE~\cite{wu2025qwenimagetechnicalreport} & F2~\cite{blackforestlabs2025flux2} & F1.K.Max~\cite{labs2025flux1kontextflowmatching} \\
    \midrule
    \textbf{mIoU} & 40.6 & 42.3 & 43.5 & 45.3 & 47.1 & 48.7 & \textbf{49.1} \\
    \bottomrule
  \end{tabular}
  }
  \vspace{-3.8em}
\end{table}

\begin{table}[h]
    \centering
    \scriptsize
    \setlength{\tabcolsep}{2pt}
    \renewcommand{\arraystretch}{0.85}
    \caption{Fusion step ablation (mIoU) across target domains using Cityscapes-pretrained Mask2Former using Flux1. Kontext Max. Base model implies no TTM.}
    \label{tab:g_fusion_ablation}
    \vspace{-0.5em}
    \begin{tabular}{@{} l cccc @{}}
        \toprule
        \textbf{Method} & \textbf{ACDC} & \textbf{Dark Zurich} & \textbf{BDDNight} & \textbf{BDD} \\
        \midrule
        Base Model       & 60.5 & 40.6 & 40.6 & 58.0 \\
        \midrule
        w/o Fusion & 60.6 & 48.7 & 46.7 & 55.7 \\
        w/ Fusion  & \textbf{64.3} & \textbf{51.7} & \textbf{49.1} & \textbf{58.7} \\
        \bottomrule
    \end{tabular}
    \vspace{-3.5em}
\end{table}

\begin{table*}[t]
\centering
\caption{Object detection performance (mAP@50) on Cityscapes→BDD100KNight-Det in a domain generalization setting.}
\label{tab:B_MainTable_Detection_Results}
\vspace{-0.5em}
\resizebox{\textwidth}{!}{
\begin{tabular}{@{}llccccccc|cc@{}}
\toprule
\multirow{2}{*}{\textbf{Model}} & \multirow{2}{*}{\textbf{Method}} & \multicolumn{7}{c|}{\textbf{Per-Class mAP@50 (\%)}} & \multicolumn{2}{c}{\textbf{Overall}} \\
\cmidrule(lr){3-9} \cmidrule(lr){10-11}
& & person & rider & car & truck & bus & motor. & bicycle & \textbf{mAP@50} & \textbf{$\Delta$$\uparrow$}
 \\
\midrule
\multirow{3}{*}{\textbf{Faster R-CNN}} 
& Without TTM & 16.1 & 22.8 & 18.6 & 4.3 & 1.0 & 3.3 & \underline{41.1} & 13.4 & 
\\
& + Flux1. Kontext Max (TTM) & \underline{36.9} & \underline{29.7} & \underline{51.4} & \underline{29.7} & 12.3 & \underline{30.6} & 36.3 & \textbf{28.4} & \textcolor{green!60!black}{\textbf{+15.0}} \\
& + QIE-2509 (TTM) & 5.0 & 22.8 & 23.7 & 10.9 & \underline{22.0} & 27.7 & \underline{34.9} & 18.4 &  \\
\midrule
\multirow{3}{*}{\textbf{Mask R-CNN}} 
& Without TTM & 15.9 & 22.8 & 13.6 & 1.3 & 0.0 & 2.5 & 25.6 & 10.2 &  \\
& + Flux1. Kontext Max (TTM) & \underline{35.1} & \underline{40.0} & \underline{51.6} & \underline{31.7} & 9.3 & 25.2 & \underline{61.1} & \textbf{31.8} & \textcolor{green!60!black}{\textbf{+21.6}} \\
& + QIE-2509 (TTM) & 4.5 & 22.8 & 24.5 & 12.1 & \underline{14.7} & \underline{36.6} & 36.6 & 19.0 &  \\
\bottomrule
\end{tabular}
}
\vspace{-0.5em}
\end{table*}

\subsection{Object Detection}

A performance comparison for the domain-generalized object detection task on Cityscapes $\rightarrow$ BDD100 K Night-Det is shown in~\cref{tab:B_MainTable_Detection_Results}. We report the mean Average Precision (mAP) metric at $0.5$ threshold. We find that TTM yields notable gains by doubling and even tripling the mAP score for Faster-RCNN and Mask R-CNN, respectively. 
Qualitative results are shown in~\cref{fig:qual_results_det}. As is visually evident from the figure, detecting objects becomes much easier after TTM, as challenging nighttime conditions are turned into easier daytime ones.

\begin{figure}[t]
\centering

\begin{minipage}[t]{0.49\linewidth}
    \captionsetup{type=table}
    \caption{Segmentation results with TTM on BDD100K-Night and ACDC using a Cityscapes pretrained Mask2Former: comparing large and small I2I models.}
    \label{tab:f_inference_time}
    \footnotesize
    \setlength{\tabcolsep}{2pt}

    \begin{adjustbox}{max width=\linewidth}
    \begin{tabular}{llcc}
        \toprule
        & \textbf{I2I Generation Model} & \textbf{mIoU$^\text{BDDNight}$} & \textbf{mIoU$^\text{ACDC}$} \\
        \midrule
        & Base Model (w/o TTM)              & 40.6          & 60.5 \\
        \midrule
        \multirow{3}{*}{\rotatebox[origin=c]{90}{\textbf{Large}}}
        & Flux.1 Kontext Dev [8 steps]      & 40.1          & 57.0 \\
        & QIE-2509                          & 47.1          & 66.4 \\
        & Flux.1 Kontext Max                & 49.1          & 64.3 \\
        \midrule
        \multirow{4}{*}{\rotatebox[origin=c]{90}{\textbf{Small}}}
        & Flux.2 Klein Base 9B [4 steps]        & \textbf{50.9} & \textbf{67.7} \\
        & Flux.2 Klein Base 4B [4 steps]    & 50.2          & 65.1 \\
        & Flux.2 Klein 9B                   & 49.9          & 67.4 \\
        & Flux.2 Klein 4B                   & \textbf{50.9} & 64.0 \\
        \bottomrule
    \end{tabular}
    \end{adjustbox}
\end{minipage}
\hfill
\begin{minipage}[t]{0.475\linewidth}
    \captionsetup{type=table}
    \caption{Image classification performance comparison on ImageNet-R dataset. Top-1 accuracies are reported.}
    \label{tab:C_imagenetR_classification_results}
    \footnotesize
    \setlength{\tabcolsep}{2pt}
    \renewcommand{\arraystretch}{0.75}
    \begin{adjustbox}{max width=\linewidth}
    \begin{tabular}{lcc}
        \toprule
        \textbf{Method} & \textbf{ImageNet-R (\%)} & \textbf{$\Delta\uparrow$} \\
        \midrule
        ResNet-50 & 36.1 & \\
        \quad + ImageNet-21K \textit{Pretraining} & 37.2 & +1.1 \\
        \quad + CBAM (\textit{Self-Attention}) & 36.8 & +0.7 \\
        \quad + $\ell_{\infty}$ Adversarial Training & 31.4 & -4.7 \\
        \quad + Speckle Noise & 37.9 & +1.8 \\
        \quad + Style Transfer Augmentation & 41.5 & +5.4 \\
        \quad + AugMix & 41.1 & +5 \\
        \quad + DeepAugment & 42.2 & +6.1 \\
        \quad + DeepAugment + AugMix & 46.8 & +10.7 \\
        \midrule
        \textbf{ResNet-50 + TTM (Ours)} & \textbf{60.8} & \textcolor{green!60!black}{\textbf{+24.7}} \\
        \midrule
        ResNet-152 & 41.3 & \\
        \textbf{ResNet-152 + TTM (Ours)} & \textbf{63.5} & \textcolor{green!60!black}{\textbf{+22.2}} \\
        \bottomrule
    \end{tabular}
    \end{adjustbox}
\end{minipage}
\vspace{-1em}   
\end{figure}

\begin{figure}[tb]
\centering
\footnotesize

\definecolor{correctgreen}{RGB}{0,150,0}
\definecolor{incorrectred}{RGB}{220,20,20}

\setlength{\tabcolsep}{1.5pt}
\renewcommand{\arraystretch}{0.98}

\newcommand{\imgheight}{1.6cm}

\begin{tabular}{@{}ccccccc@{}}
\includegraphics[height=\imgheight, width=0.132\columnwidth, keepaspectratio=false]{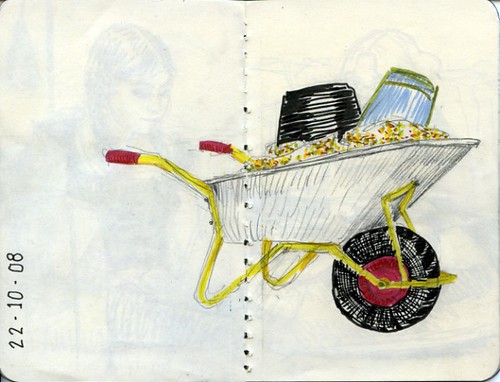} &
\includegraphics[height=\imgheight, width=0.132\columnwidth, keepaspectratio=false]{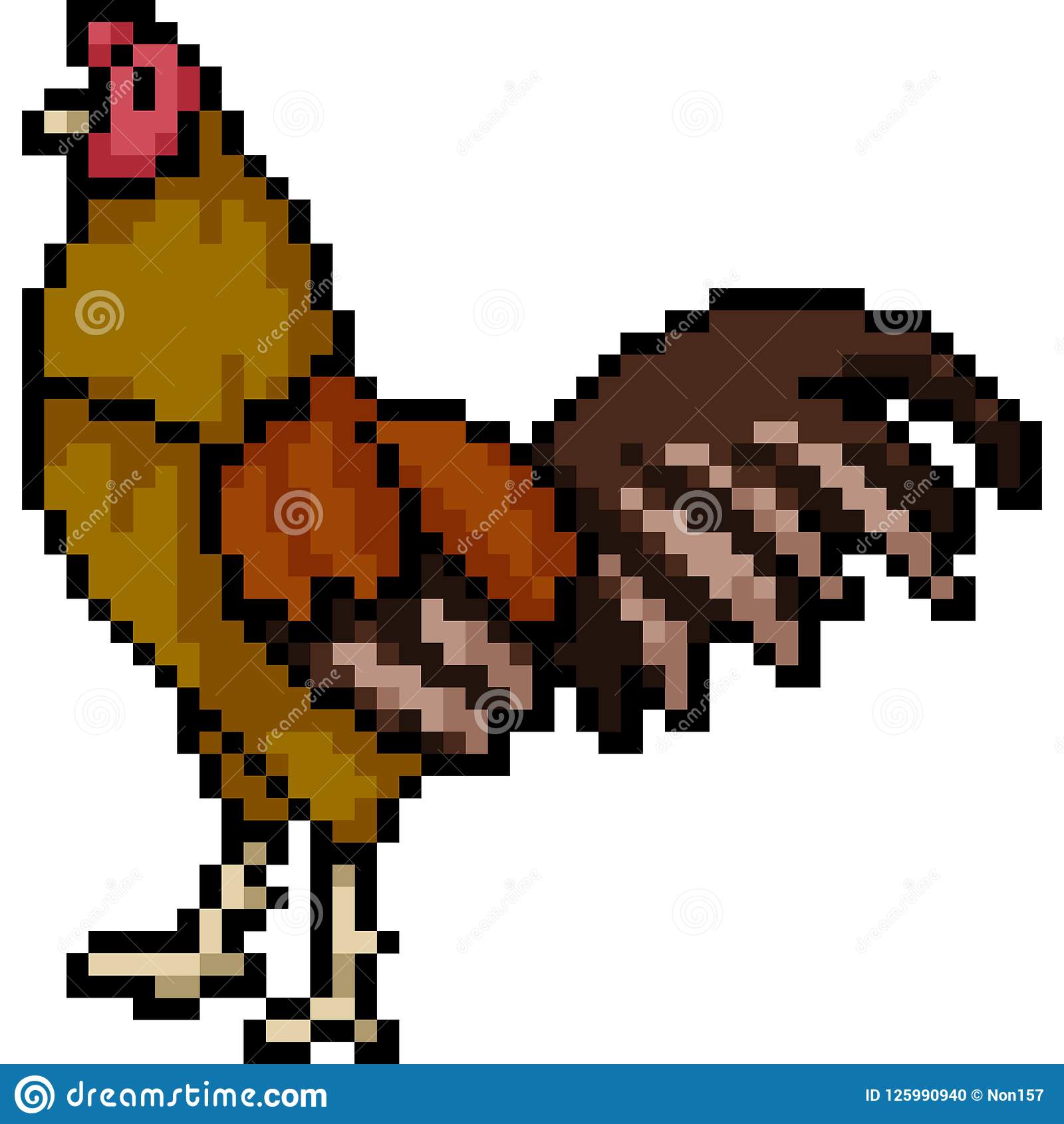} &
\includegraphics[height=\imgheight, keepaspectratio=false]{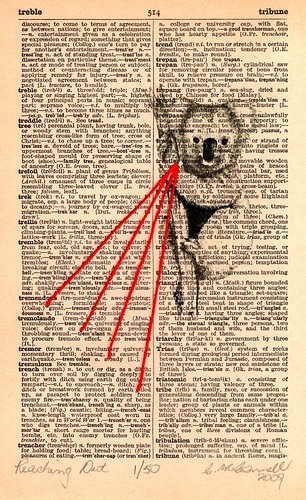} &
\includegraphics[height=\imgheight, width=0.132\columnwidth, keepaspectratio=false]{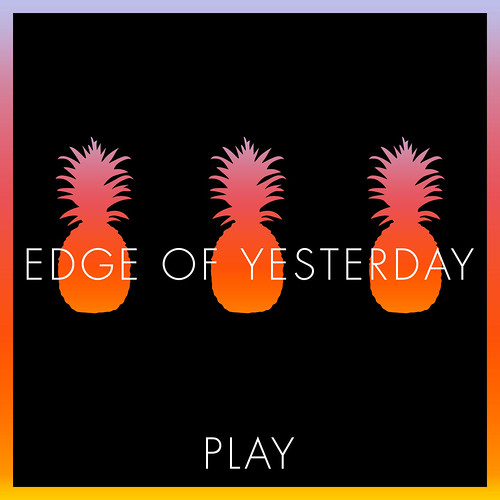} &
\includegraphics[height=\imgheight, width=0.132\columnwidth, keepaspectratio=false]{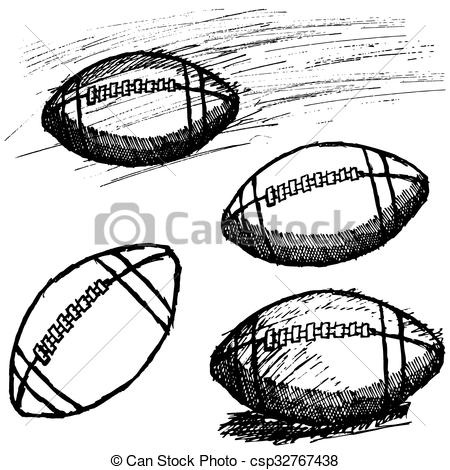} &
\includegraphics[height=\imgheight, width=0.132\columnwidth, keepaspectratio=false]{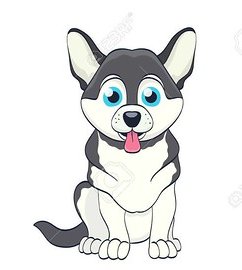} &
\includegraphics[height=\imgheight, width=0.132\columnwidth, keepaspectratio=false]{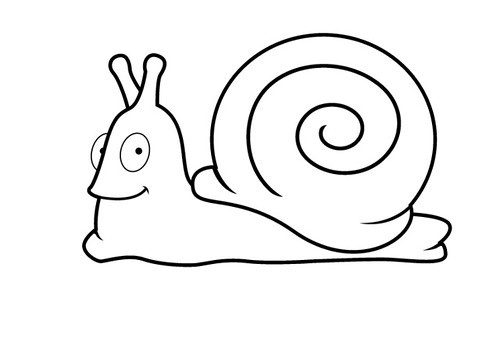}
\\
\scriptsize\textcolor{incorrectred}{broom} &
\scriptsize\textcolor{incorrectred}{stole} &
\scriptsize\textcolor{incorrectred}{beacon} &
\scriptsize\textcolor{incorrectred}{ant} &
\scriptsize\textcolor{incorrectred}{tennis ball} &
\scriptsize\textcolor{incorrectred}{Chihuahua} &
\scriptsize\textcolor{incorrectred}{joystick}
\\[0.3em]
\multicolumn{7}{c}{\scriptsize$\Downarrow$ \hspace{0.5em}\textbf{w/ TTM}}
\\[0.2em]
\includegraphics[height=\imgheight, width=0.132\columnwidth, keepaspectratio=false]{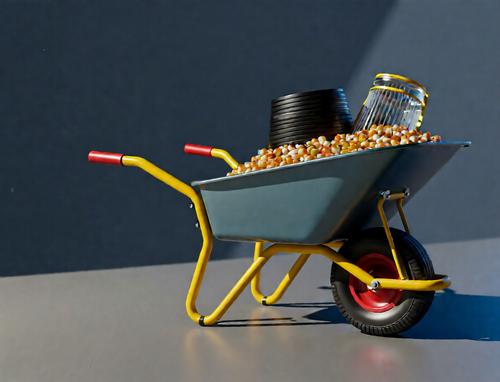} &
\includegraphics[height=\imgheight, width=0.132\columnwidth, keepaspectratio=false]{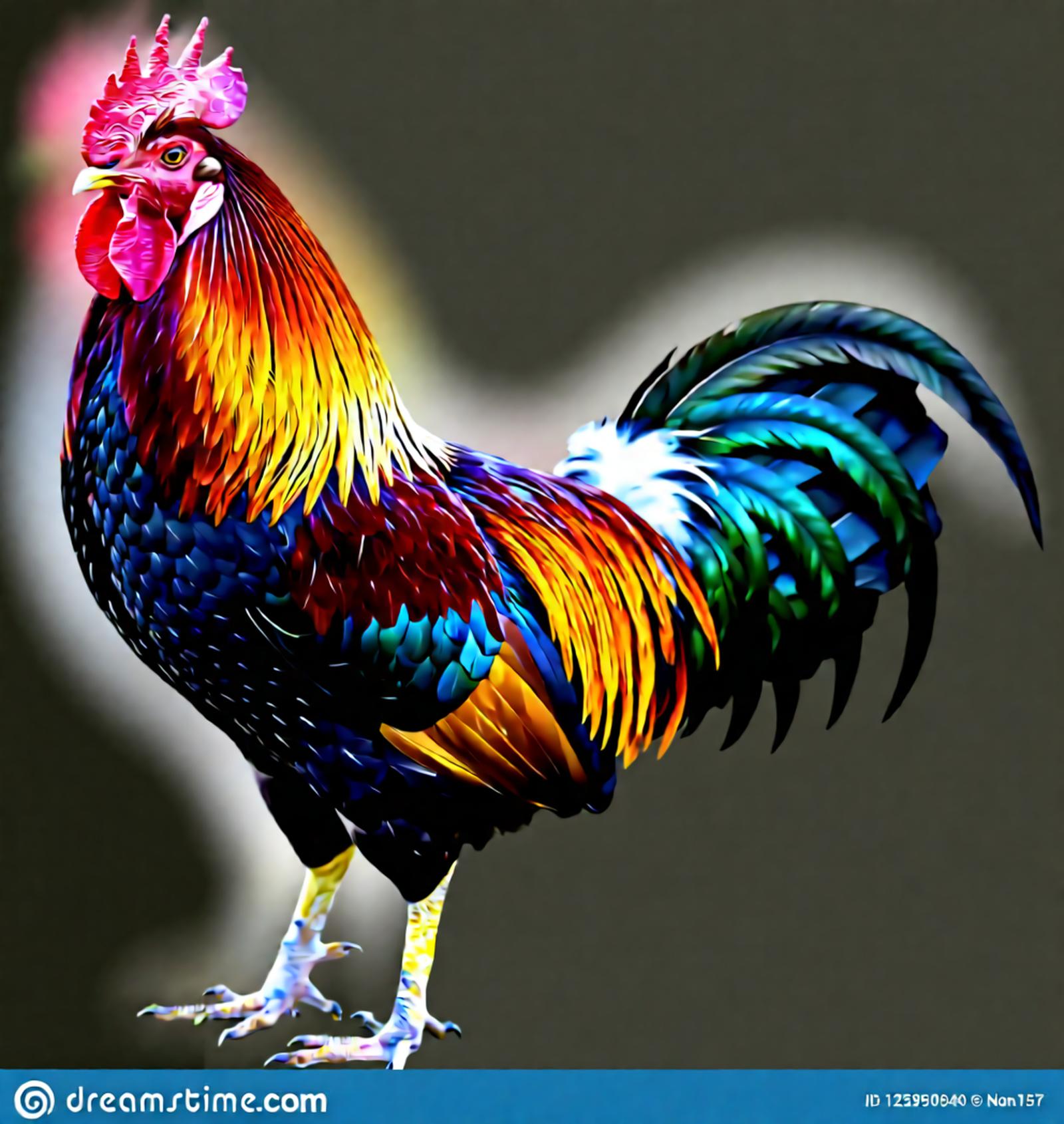} &
\includegraphics[height=\imgheight, keepaspectratio=false]{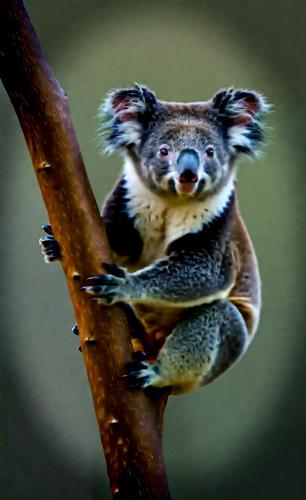} &
\includegraphics[height=\imgheight, width=0.132\columnwidth, keepaspectratio=false]{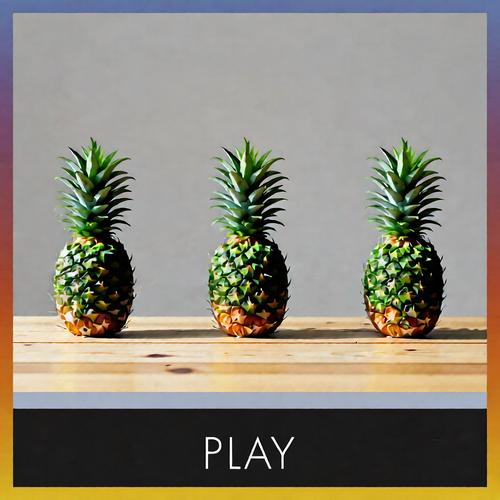} &
\includegraphics[height=\imgheight, width=0.132\columnwidth, keepaspectratio=false]{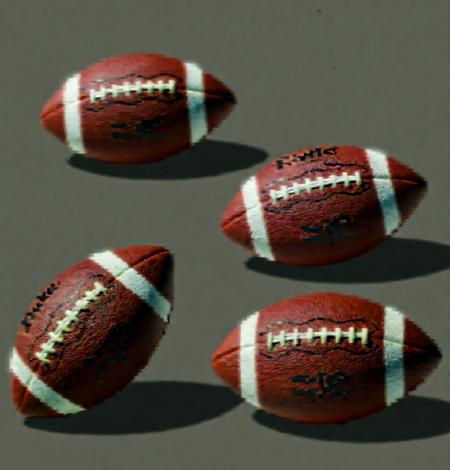} &
\includegraphics[height=\imgheight, width=0.132\columnwidth, keepaspectratio=false]{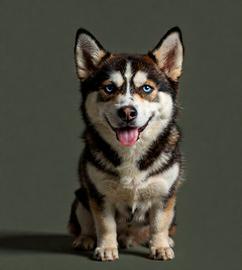} &
\includegraphics[height=\imgheight, width=0.132\columnwidth, keepaspectratio=false]{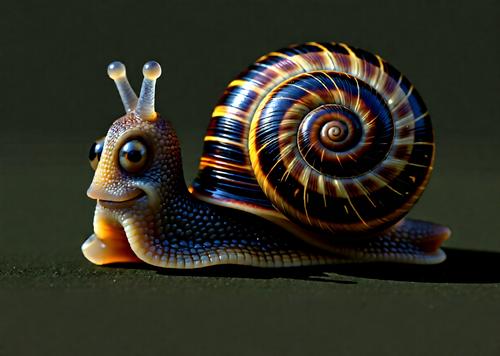}
\\
\scriptsize\textcolor{correctgreen}{barrow} &
\scriptsize\textcolor{correctgreen}{hen} &
\scriptsize\textcolor{correctgreen}{koala} &
\scriptsize\textcolor{correctgreen}{pineapple} &
\scriptsize\textcolor{correctgreen}{rugby ball} &
\scriptsize\textcolor{correctgreen}{siberian husky} &
\scriptsize\textcolor{correctgreen}{snail}
\\
\end{tabular}

\caption{
Performance comparison on domain generalized image classification with ResNet-50.
Top row: predictions on original ImageNet-R images (without TTM).
Bottom row: predictions on the same images after TTM.
\textcolor{correctgreen}{Green} = correct, \textcolor{incorrectred}{red} = incorrect.
}
\vspace{-1.5em}
\label{fig:classification_qual_results}
\end{figure}

\vspace{-1.0em}   

\subsection{Image Classification}
\label{subsec:classification_exp}
Beyond our main semantic segmentation and object detection tasks, we additionally analyze TTM on an image classification task to assess its behavior in a substantially different setting.
For image classification task, apart from the base model (w/o TTM), we also compare against other models introduced in~\cite{hendrycks2021many} that are trained on ImageNet-1K while using additional strategies to reduce the domain gap like pre-training on ImageNet-21K (10x the size of ImageNet-1K), different augmentation techniques, using self-attention for better spatial feature learning, and adversarial training. However, we only take raw ResNet-50 and ResNet-152 models pre-trained on ImageNet-1K as candidates for applying TTM without using any of the additional domain reduction techniques. Here, due to the large validation set size, for image generation, we use a 4-bit quantized variant of QIE-2509 with 8-step Lightning LoRA weights for faster inference. The improvement should be even greater if we use the non-quantized complete QIE-2509 model.
 
Quantitative results with our TTM method are shown in~\cref{tab:C_imagenetR_classification_results}. As can be seen, we are able to get the highest top-1 accuracy with ResNet-50 using TTM on ImageNet-R without employing any additional training strategies. Moreover, even with the larger ResNet-152 model, we are able to significantly improve the top-1 accuracy. Additionally, following a similar trend to domain generalized semantic segmentation, applying TTM with ResNet-50 even surpasses the performance of a larger ResNet-152 model (from 41.3 to 60.8). Qualitative results are shown in~\cref{fig:classification_qual_results}. As visible from the figure, the transformation of different renditions to real-life-like objects reduces the domain gap w.r.t ImageNet thereby helping to achieve better image classification performance for a fixed discriminative model. 

\vspace{-1em}   

\subsubsection{\textbf{\textit{Inference Time and Deployment.}}}
\label{para:deployment}
TTM introduces a modular image transformation step that operates independently of the downstream perception model. 
We analyze the runtime characteristics of the TTM pipeline, focusing on two components: prompt generation via MLLM metaprompting and the inference speed of the I2I generation model.

\textit{MLLM metaprompting:} We employ a single task-level prompt shared across all images. As a result, the MLLM is invoked only once to generate the source-domain prompt, and this step is detached from the per-image inference pipeline.

\textit{I2I generation:} The main computational step of TTM is the I2I transformation applied to each test image. Recent advances in efficient generative architectures substantially reduce runtime and memory requirements while maintaining high generation quality. \Cref{tab:f_inference_time} compares the post-TTM segmentation performance of lightweight and large I2I models on the BDDNight and ACDC datasets. The results show that reducing model complexity does not compromise downstream performance.

\begin{figure}[tb]
  \centering
  \includegraphics[height=6.5cm]{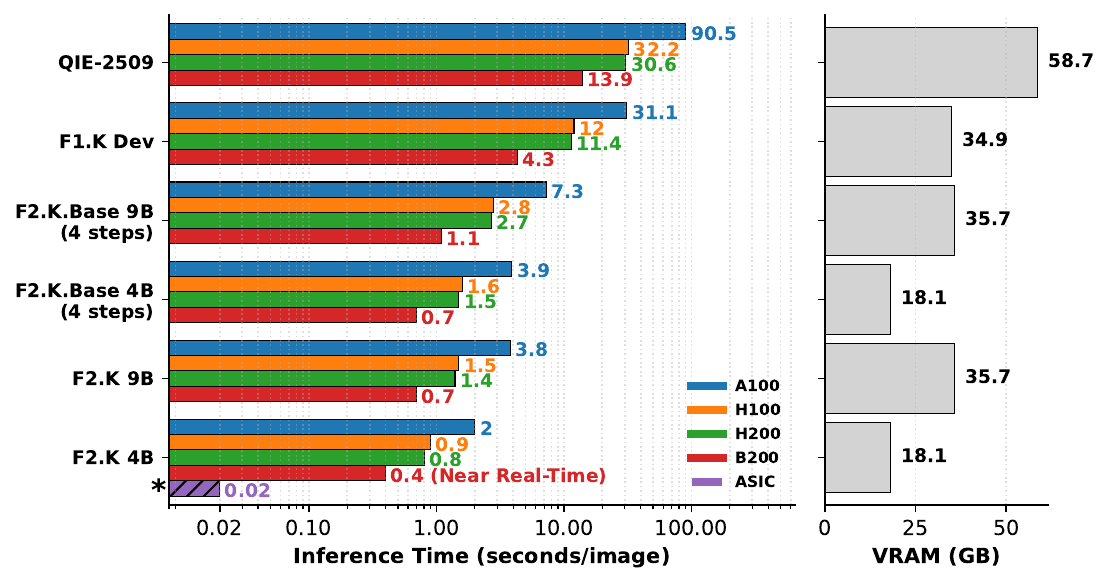}
  \caption{Average generation time per image for different I2I generation models across different GPUs in seconds per image. VRAM occupancy for each model is also depicted on the right. $*$ bar refers to ASIC chip~\cite{Groq,Cerebras} inference time estimates using public benchmarks. Note that we run Flux.2 Klein Base (4B/9B) for just 4 steps, as this already produces satisfactory images for TTM. We already reach near real-time inference with smaller models.}
  \label{fig:inference_time_comp_plot}
  \vspace{-1.5em}
\end{figure}

In particular, the distilled Flux.2 Klein 4B model achieves higher mIoU on BDDNight than larger models such as QIE-2509 while providing an $\sim$45$\times$ speedup (\cref{fig:inference_time_comp_plot}) on an A100 GPU (90.5 vs.\ 2s/img). Its performance on ACDC remains comparable to that of larger models. Similarly, Flux.2 Klein Base 9B with only four inference steps outperforms larger models on both BDDNight and ACDC while reducing latency by $\sim$12$\times$ (90.5 vs.\ 7.3s/img).

As illustrated in \cref{fig:inference_time_comp_plot}, modern accelerators further accelerate this process through higher throughput and low-precision execution. For example, Flux.2 Klein 4B generates an image in 2s on an A100 GPU, 0.9s on an H100, 0.8s on an H200, and 0.4s on a B200 GPU. These results demonstrate that TTM can operate at practical inference speeds on modern hardware. Additional improvements through batching, quantization, and emerging inference accelerators (e.g., Groq LPU~\cite{Groq}, Cerebras WSE~\cite{Cerebras}) are expected to further reduce latency. Additional runtime analysis of the I2I generation models within the TTM framework is included in the appendix.

\vspace{-0.5em}

\section{Conclusion}
\label{sec:conclusion}
\vspace{-0.5em}   
In this paper, we propose an efficient approach to improve the perception robustness during inference time called Test-Time Modification (TTM). During TTM, we perform an inverse domain transformation of the target domain images back to the source domain by implicitly distilling the world knowledge of state-of-the-art Image-to-Image generation models. To demonstrate the effectiveness of our approach, we experimentally focus on the domain generalization setting for semantic segmentation, object detection, and image classification tasks, utilizing challenging target data distributions across various weather and daytime conditions, as well as rendering shifts. 
The significant improvements in our quantitative and qualitative results across multiple downstream tasks, datasets, and generation models demonstrate that test-time modification using large, information-rich Image-to-Image generation models can significantly help in creating more robust perception pipelines for challenging environments.
Finally, our deployment analysis shows that these gains are achievable with %
realistic latencies today, and that continued advances in efficient architectures and accelerator hardware make real-time deployment increasingly feasible. %

\section*{Acknowledgements}
The research is funded by the German Federal Ministry for Economic Affairs and Energy within the project “NXT GEN AI METHODS – Generative Methoden für Perzeption, Prädiktion und Planung". We thank the consortium and also acknowledge the Gauss Centre for Supercomputing for providing computing time on the GCS Supercomputer JUWELS at Jülich Supercomputing Centre (JSC).

\bibliographystyle{splncs04}
\bibliography{main}
\end{document}